\documentclass[11pt]{article}

\PassOptionsToPackage{table}{xcolor}
\usepackage[preprint]{acl}
\usepackage{times}
\usepackage{latexsym}
\usepackage[T1]{fontenc}
\usepackage[utf8]{inputenc}
\usepackage{microtype}
\usepackage{inconsolata}
\usepackage{booktabs}
\usepackage{amsfonts}
\usepackage{nicefrac}
\usepackage{amsmath}
\usepackage{amsthm}
\usepackage{algorithm}
\usepackage{algpseudocode}
\usepackage{graphicx}
\usepackage{multirow}
\usepackage{array}
\usepackage{makecell}
\newtheorem{theorem}{Theorem}

\title{Information Gain-based Rollout Policy Optimization: An Adaptive Tree-Structured Rollout Approach for Multi-Turn Search Agents}

\author{
\textbf{Yijun Zhang}$^{\ast}$ \quad
\textbf{Fan Xu}$^{\ast}$ \quad
\textbf{Jiaxin Ding}$^{\dagger}$ \quad
\textbf{Yule Xie} \\
\textbf{Xin Ding} \quad
\textbf{Haoxiang Zhang} \\
Shanghai Jiao Tong University \\
{\small $^{\ast}$Equal contribution. \quad
$^{\dagger}$Corresponding author.}
}

\begin{document}

\maketitle


\begin{abstract}
    Reinforcement learning (RL) improves large language model (LLM) agents on long-horizon search tasks that require multiple intermediate decisions before a final outcome. However, rollout budgets are often allocated without assessing intermediate-state utility, which can waste computation on unpromising branches. We propose Information Gain-based Rollout Policy Optimization (IGRPO), a framework that organizes rollout collection around intermediate-state informativeness. Specifically, IGRPO performs budget-aware tree-structured rollouts in which expansion probabilities depend on node-level informativeness, allowing informative branches to receive more computation while less informative branches are expanded less frequently within a fixed rollout budget. By directly shaping how training trajectories are generated, IGRPO induces a limiting teacher distribution over search trajectories that favors higher cumulative informativeness. The resulting distribution provides an explicit policy optimization target, connecting adaptive rollout collection with principled policy learning.
    Experiments on seven search-augmented question answering benchmarks show that IGRPO achieves higher average accuracy than strong baselines on both 3B and 7B backbones under comparable rollout budgets, supporting informativeness-guided trajectory generation for training search agents.
    Code is available at \url{https://github.com/e3trange/IGRPO}.
\end{abstract}

\section{Introduction}
Large language models (LLMs)~\cite{brown2020language, touvron2023llama, chowdhery2023palm} are increasingly trained to act as agents that solve tasks through multi-turn interaction with external tools~\cite{yao2022react, schick2023toolformer, zhang2025landscape, huang2025deep}. In search-augmented question answering~\cite{wang2024searching, li2025search, jin2025search}, an agent must decide what to reason about, when to issue a retrieval query, how to use the evidence, and finally when to answer. A central difficulty in this setting is that the agent must allocate a limited interaction budget across many possible intermediate states. Effective training therefore requires not only assigning credit to completed trajectories, but also deciding where rollout computation should be spent.

Reinforcement learning has been widely adopted to improve such agents by optimizing their interaction trajectories~\cite{qian2025toolrl}. Existing outcome-based methods typically optimize trajectories using final correctness rewards~\cite{schulman2017proximal, ouyang2022training}, while group-based variants such as GRPO~\cite{shao2024deepseekmath} further avoid a learned critic by comparing multiple rollouts for the same question.
Recent studies have sought to improve long-horizon agent training from two complementary directions: finer-grained credit assignment and broader exploration. Chain-based methods such as GiGPO~\cite{feng2025group}, HGPO~\cite{he2026hierarchyofgroups}, and IGPO~\cite{wang2025information} move beyond pure outcome-level learning through finer-grained credit assignment for intermediate decisions. Tree-based methods such as AEPO~\cite{dong2025aepo} and Tree-GRPO~\cite{ji2025tree} extend rollout generation from a single chain to a branching search process, exploring alternative continuations through random branching or heuristic hard-threshold expansion rules. Together, these two lines of work demonstrate that intermediate states are crucial for training search agents.

\begin{figure*}[t]
      \centering
      \includegraphics[width=\linewidth]{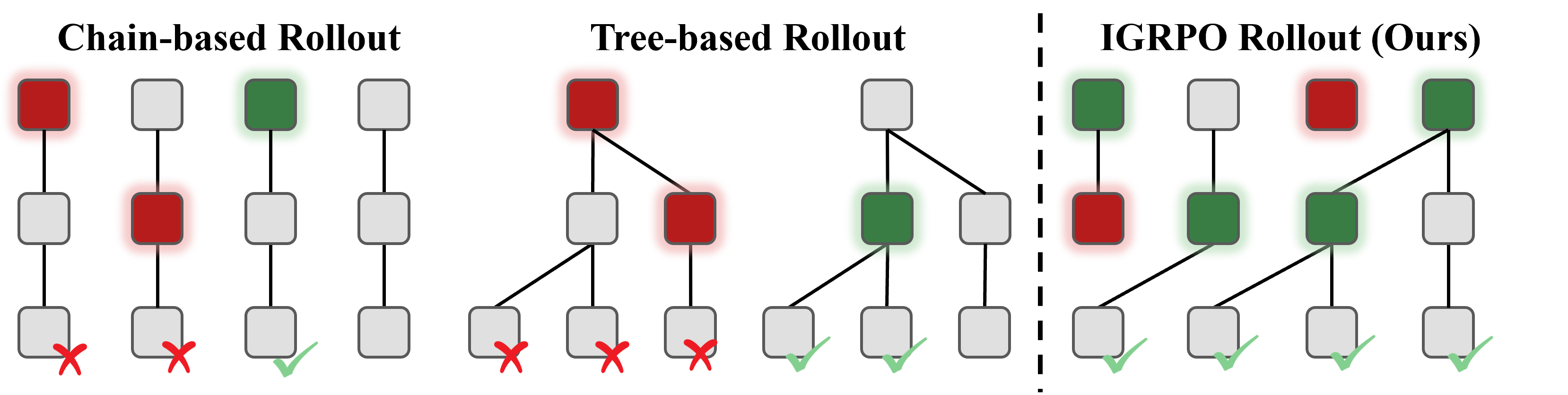}
      \caption{Illustration of different patterns. Red nodes represent unpromising states, green nodes represent informative states, and gray nodes represent other intermediate states. \textbf{Left}: both chain-based and existing tree-based methods may still allocate exploration budget to unpromising nodes. \textbf{Right}: IGRPO adopts information gain-based rollout allocation, preferentially expanding informative nodes and reducing exploration over unpromising branches.}
      \label{fig:intro_rollout}
\end{figure*}

\textit{However, these methods do not directly assess whether further node expansion aids task completion, leaving trajectory generation insufficiently guided by node utility.} Chain-based methods may spend full trajectories on prefixes that are already unpromising. Tree-based methods improve exploration through branching, but their heuristic or uncertainty-driven expansion rules can still allocate branches to unpromising continuations.
Moreover, these methods typically treat intermediate nodes as expansion targets rather than explicitly assessing whether they provide useful evidence for the final answer. As a result, training can waste computation on low-value or even misleading prefixes while under-exploring states that contain evidence useful for answering the question. This limitation appears in both chain-based and existing tree-based rollouts, as illustrated in the left panel of Figure~\ref{fig:intro_rollout}.

To tackle this issue, we propose \textbf{I}nformation \textbf{G}ain-based
\textbf{R}ollout \textbf{P}olicy \textbf{O}ptimization (IGRPO), a policy optimization framework that treats
intermediate state informativeness as the core organizing principle for rollout collection.
We use ``information gain'' in the operational sense of \emph{answer-likelihood information gain}: the increase in a policy's normalized likelihood of the ground-truth answer after reaching an intermediate search state. Instead of allocating search budget uniformly over active branches, IGRPO treats information gain-derived informativeness as a soft expansion potential. At each rollout stage, active prefixes with larger informativeness are selected for expansion with higher probability, while low-informativeness prefixes are sampled less often and therefore consume less computation. This induces a budget-aware tree-structured rollout process that is better aligned with the needs of long-horizon search (see Figure~\ref{fig:intro_rollout}, right). Moreover, our theoretical analysis shows that this information gain-based rollout induces a limiting teacher distribution that explicitly biases sampling toward more informative trajectories. This in turn provides a clear policy optimization target, enabling the policy to be trained toward the induced distribution. In this way, IGRPO unifies adaptive
tree-structured rollout collection and policy optimization within a single framework for search-intensive LLM agents. Our main contributions are summarized as follows:

\begin{itemize}
      \item We propose information gain-based tree rollouts that adaptively allocate a fixed budget to informative intermediate states.

      \item We characterize the limiting rollout distribution, establishing a teacher distribution as the target for policy optimization.

      \item Across challenging QA benchmarks, IGRPO achieves average accuracy gains of 2.9\% (3B) and 1.8\% (7B) over the strongest baseline.
\end{itemize}

\section{Related Work}
\begin{figure*}[t]
    \centering
    \includegraphics[width=\textwidth]{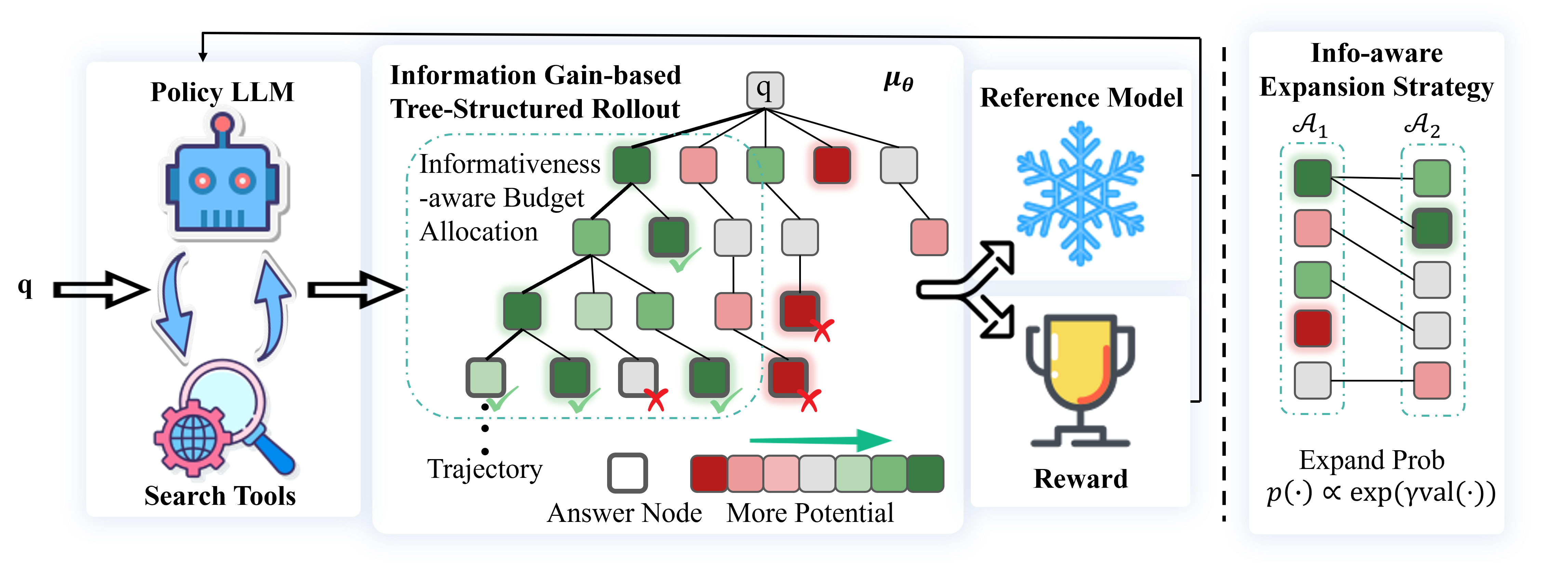}
    \caption{Overview of IGRPO. \textbf{Left:} During rollout, the expansion budget is adaptively allocated according to node-level informativeness, so that more informative nodes are expanded with higher probability, thereby inducing a distribution $\mu_{\theta}$ that biases sampling toward more promising trajectories. \textbf{Right:} the expansion process between tree-level 1 and tree-level 2 is governed by an informativeness-aware strategy, where each active node is selected for expansion with probability
        $p(\cdot)\propto \exp(\gamma\, \text{val}(\cdot))$.
    }
    \label{fig:pipeline}
\end{figure*}

\paragraph{RL for LLMs.}
RL improves LLM alignment and reasoning~\cite{ouyang2022training, guo2025deepseek}. PPO~\cite{schulman2017proximal} optimizes policies with a learned critic, whereas GRPO~\cite{shao2024deepseekmath}, RLOO~\cite{ahmadian2024back}, DAPO~\cite{yu2025dapo}, and GSPO~\cite{zheng2025group} estimate advantages without an explicit value model.

\paragraph{RL for LLM search agents.}
Early works establish RL frameworks for interleaving reasoning and search across multiple interaction turns. Search-R1~\cite{jin2025search} and R1-Searcher~\cite{song2025r1} train search policies with outcome-based rewards. ZeroSearch~\cite{sun2025zerosearch} uses simulated retrieval during training, while Smart-Searcher~\cite{song2025smart} combines internal knowledge with retrieved evidence.

To alleviate the sparsity of outcome-only rewards in multi-turn search, several works move toward
finer-grained supervision over intermediate decisions. StepSearch~\cite{wang2025stepsearch} introduces turn-wise PPO with richer intermediate rewards for evidence utilization. GiGPO~\cite{feng2025group} and HGPO~\cite{he2026hierarchyofgroups} group steps by shared states and historical contexts, respectively. IGPO~\cite{wang2025information} uses information gain as an intrinsic turn-level reward.

Another direction extends rollout generation from a single chain to a branching search process. Tree-GRPO~\cite{ji2025tree} explores alternative continuations through random tree-structured rollouts, while ARPO~\cite{dong2025agentic} and AEPO~\cite{dong2025aepo} guide branching with entropy. However, these methods may still allocate exploration budget to unpromising branches and do not explicitly aim to shape the trajectory sampling distribution. IGRPO instead uses the information gain signal to favor informative trajectories under a fixed rollout budget, with the induced teacher distribution guiding policy optimization, thus explicitly targeting the generation of higher-quality trajectories for policy training.

\section{Preliminaries}
\subsection{Problem Setup}

Let $\mathcal{D}=\{(q,a)\}$ denote a dataset of question-answer pairs, where $q$ is a question and $a$ is the ground-truth answer. We consider a search environment equipped with an external retrieval tool, denoted by $\mathcal{S}$, which accepts a textual query and returns a set of retrieved results, such as documents. The goal of the agent is to solve question $q$ by interacting with $\mathcal{S}$ under a limited search budget and finally producing an answer $\hat{a}$. For each question $q$, the agent generates a search rollout $o=(\tau_1,\tau_2,\dots,\tau_T)$, where $T$ denotes the total number of turns and each $\tau_i$ is a particular interaction turn. The last turn $\tau_T$ is the answer turn, which outputs the final prediction $\hat{a}$ in a $[\textsc{ans}]$ step, while each preceding turn corresponds to one search turn. In particular, for $t<T$, each turn $\tau_t$ is defined as a triple consisting of $[\textsc{think}]$, $[\textsc{search}]$, and $[\textsc{result}]$ steps. The $[\textsc{think}]$ step represents the intermediate reasoning based on the current search history $h_{t-1}$, which consists of the previous search-result pairs collected up to turn $t-1$. The $[\textsc{search}]$ step issues a textual retrieval query to the external search tool $\mathcal{S}$. The $[\textsc{result}]$ step then returns the retrieved evidence from $\mathcal{S}$. We then define the reward of the rollout $o$ as $R(o)=\mathbb{I}[\hat{a}=a]$, where $\mathbb{I}[\cdot]$ is the indicator function. The objective is to maximize the average reward of sampled rollouts.

\subsection{Group-based RL Pipeline}

Group Relative Policy Optimization (GRPO)~\cite{shao2024deepseekmath} trains search agents toward the following KL-regularized expected-reward objective:
\begin{equation}
    \begin{aligned}
         & \max_{\pi_\theta}\;
        \mathbb{E}_{(q,a)\sim\mathcal{D}}
        \left[
            \mathbb{E}_{o\sim \pi_\theta(\cdot\mid q)}[R(o)]
            \right]                       \\
         & \quad-\beta\,\mathbb{D}_{\mathrm{KL}}(\pi_\theta\|\pi_{\mathrm{ref}}).
    \end{aligned}
    \label{eq:grpo-regularized}
\end{equation}
where $\pi_\theta$ is the actor policy, $\pi_{\mathrm{ref}}$ is the reference policy, and $\beta$ controls the KL penalty.

For each question $q$, GRPO independently samples $N$ complete rollouts from the frozen policy $\pi_{\theta_{\mathrm{old}}}(\cdot\mid q)$. Under a tree representation, each rollout $o_i$ corresponds to a root-to-leaf path: partial search histories $\{h_{i,t}\}_{t=0}^{T_i-1}$ constitute nonterminal nodes, and interaction turns $\{\tau_{i,t}\}_{t=1}^{T_i}$ form edges, with the final answer turns $\tau_{i,T_i}$ leading to terminal nodes $\{h_{i,T_i}\}$. Each path starts from $h_{i,0}$, which contains only the prompt and question, and is generated without reusing intermediate nodes. GRPO computes an advantage for each path by normalizing rewards within the group, then applies PPO-style clipped updates~\cite{schulman2017proximal} to generated tokens while masking retrieved tokens.

\section{Methodology}
In this section, we describe our methodology. An overview of how the pipeline reshapes the sampling distribution to favor informative training trajectories is illustrated in Figure~\ref{fig:pipeline}.

\subsection{Tree-Structured Rollout Framework}
\label{sec:rollout}

\begin{figure}[t]
    \centering
    \includegraphics[width=\linewidth]{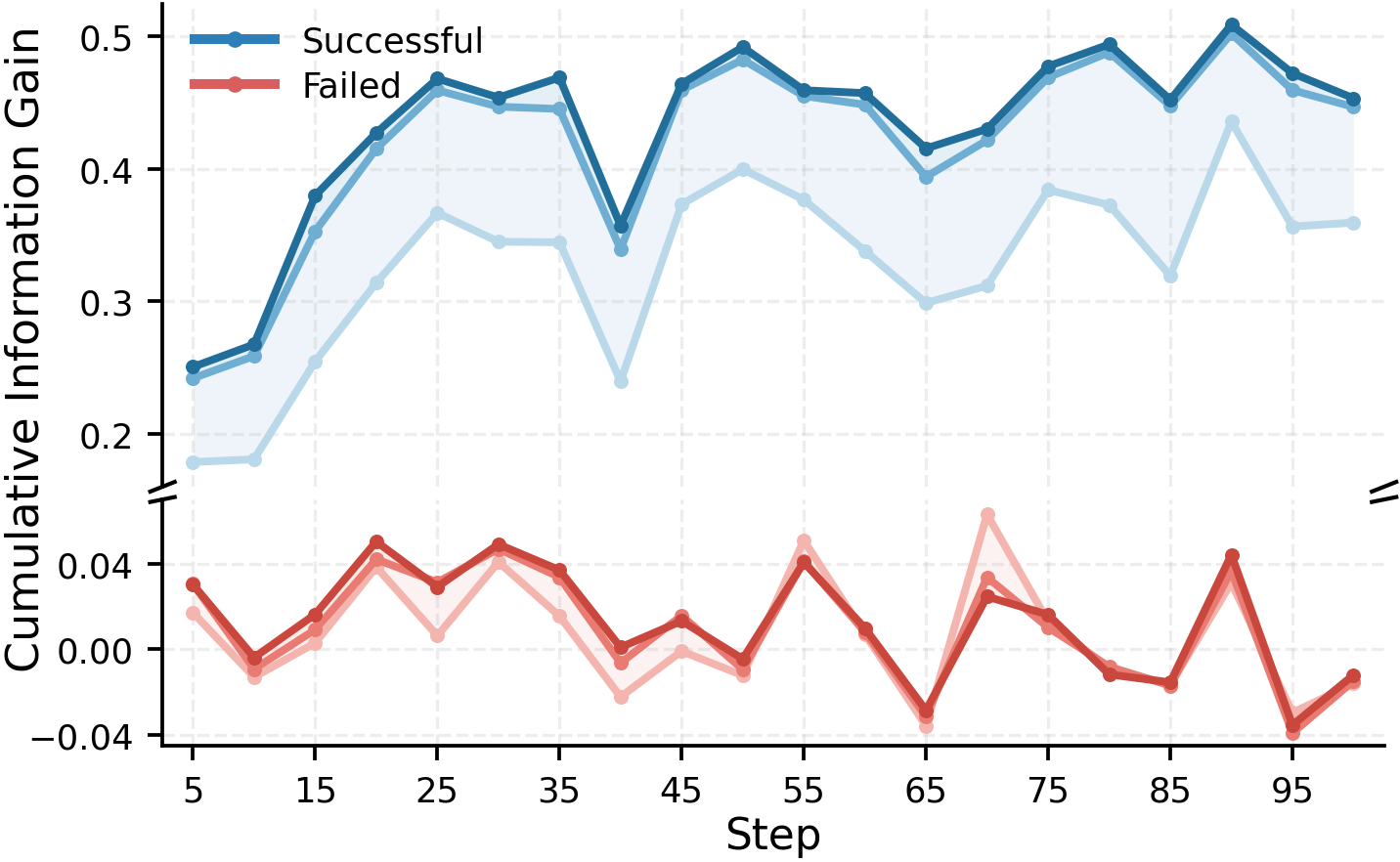}
    \caption{Average cumulative information gain at different trajectory turns (start from turn 1). Blue and red denote successful and failed trajectory groups, respectively, and color intensity corresponds to the search turn.}
    \label{fig:pre}
\end{figure}

\paragraph{Node-level Information Gain.} To instantiate the idea of tree expansion, we first require a criterion for assessing how promising each intermediate node is. We employ information gain-derived value as a scoring signal for intermediate search nodes, a criterion further supported by a preliminary empirical observation. Specifically, during IGPO~\cite{wang2025information} training, we compare the average cumulative information gain of sampled trajectories that eventually answer the question correctly with those that fail. As shown in Figure~\ref{fig:pre}, the gap between the two groups is substantial: successful trajectories consistently exhibit much larger average information gain. This evidence supports informativeness as a criterion for assessing node promise even before the final outcome is observed, motivating its use for rollout guidance in IGRPO rather than for reward shaping as in IGPO.

We now define the information gain. For an intermediate history node $h$, we first define its answer score under the current policy $\pi_\theta$ as the likelihood of the ground-truth answer conditioned on $h$:
\begin{equation}
    \label{eq:score}
    s_\theta(h; a)\mathrel{:=}
    \exp{\bigg(\frac{1}{L}\sum_{j=1}^L \log \pi_\theta(a_j \mid h, a_{<j})\bigg)},
\end{equation}
where $a=(a_1,\dots,a_L)$ denotes the ground-truth answer tokens. Let $\mathrm{parent}(h)$ denote the parent node of $h$ in the search tree. We then define the information gain of an intermediate node $h$ as the marginal improvement in answer score induced by expanding from its parent to the current node:
\begin{equation}
    \mathrm{IG}(h)
    =
    s_\theta(h; a)
    -
    s_\theta(\mathrm{parent}(h); a);
    \label{eq:IG}
\end{equation}
while we set $\mathrm{IG}(h_0)=s_\theta(h_0; a)$ for the root node. In this sense, the answer score can be viewed as a cumulative information gain signal along the path from the root to the current node, and the difference between two nodes' answer scores reflects the difference in cumulative information gain accumulated after they branch from their lowest common ancestor. Following previous works~\cite{jin2025search, wang2025information}, the ground-truth answer $a$ is wrapped in the same schema as a predicted answer to ensure consistency with formatting, e.g., \texttt{<think>Now there's enough information to answer.</think> <answer>Ground Truth $a$</answer>}. This allows the answer score to be computed efficiently, without any further sampling.

\paragraph{Rollout Framework.} We next utilize information gain to construct rollout trajectories under a fixed search budget. For each question $q$, we initialize a search tree with a root history node $h_0$, and define the initial active node set as $\mathcal{A}_0=\{h_0\}$, where an active node refers to an intermediate node that remains eligible for further expansion.

At the $i$-th expansion stage, let $\mathcal{A}_i$ denote the current set of active nodes. Instead of allocating rollout budget uniformly across all active nodes, we distribute it according to their relative informativeness. Concretely, for each node $h\in\mathcal{A}_i$, we assign an expansion probability
\begin{equation}
    \label{eq:expand_prob}
    p_i(h)
    =
    \frac{\exp\!\big(\gamma\, \mathrm{val}(h)\big)}
    {\sum\limits_{h' \in \mathcal{A}_i}\exp\!\big(\gamma\, \mathrm{val}(h')\big)},
\end{equation}
where $\mathrm{val(\cdot)}$ can be any IG-derived signal, and $\gamma$ is a temperature parameter controlling how strongly the allocation favors nodes with larger values. In our implementation, we use the signal defined later in~\eqref{eq:val}. Given a stage budget $B_i$, we independently sample $B_i$ active nodes from $\mathcal{A}_i$ according to the allocation probability in \eqref{eq:expand_prob}, and expand each selected node under the frozen rollout policy. The resulting intermediate nodes form the next active set $\mathcal{A}_{i+1}$, while terminal nodes receive rewards according to their final predicted answers and are then recorded as completed rollout endpoints. This process is repeated until the active set becomes empty or the maximum interaction turn is reached.

A key distinction of our rollout framework is that branching is not determined by hard thresholding. Instead, expansion is governed by a probabilistic allocation, which softly biases computation toward nodes with higher informativeness. As a result, trajectories containing more informative intermediate nodes are sampled with higher probability, thereby inducing a non-uniform rollout distribution biased by informativeness. We analyze the limiting distribution and its properties in the next subsection.

\subsection{Theoretical Analysis and Policy Objective}
\label{sec:analysis}

According to the rollout procedure described above, we first characterize the distribution of sampled trajectories induced by our information gain-based expansion strategy. Intuitively, since the expansion budget at each layer is allocated probabilistically according to node-level informativeness, trajectories passing through more informative intermediate nodes are expanded more frequently and therefore appear with higher probability in the final sample set. The following theorem formalizes the limiting trajectory distribution when the per-layer budget is sufficiently large. Without loss of generality, we continue rolling out even after an answer has been produced. As a result, every trajectory is padded to the same maximum interaction length $T$. Then, the limiting rollout distribution can be viewed as an exponentially tilted version of the original policy-induced trajectory distribution.

\begin{theorem}[Limiting trajectory distribution induced by IG-based rollout]
    \label{theorem:1}
    Consider the following tree-structured rollout process. For each question $q$,
    the process starts from the root history node $h_0=q$ and initializes the active
    set as $\mathcal{A}_0=\{h_0\}$. At the $t$-th expansion stage
    $(0\leq t<T)$, each node $h\in\mathcal{A}_t$ is assigned the expansion
    probability
    \begin{equation}
        \label{eq:expand_prob_theorem}
        p_t(h)
        =
        \frac{\exp\!\big(\gamma\,\mathrm{val}(h)\big)}
        {\sum_{h'\in\mathcal{A}_t}\exp\!\big(\gamma\,\mathrm{val}(h')\big)},
    \end{equation}
    where $\mathrm{val}(\cdot)$ denotes the informativeness of a history node. Given the stage budget $B_t$, the process independently samples $B_t$ nodes
    from $\mathcal{A}_t$ according to $p_t(\cdot)$, expands each selected node under
    the policy $\pi_\theta$, and adds the resulting child nodes to
    $\mathcal{A}_{t+1}$. For convenience, define the cumulative value of a trajectory $o = (h_0 \xrightarrow{\tau_1} h_1 \xrightarrow{\tau_2} \cdots \xrightarrow{\tau_T} h_T )$ as
    $V(o) := \sum_{t=1}^{T-1} \mathrm{val}(h_t)$. Let $\pi_\theta(o)=\prod_{t=1}^{T}\pi_\theta(\tau_t\mid h_{t-1})$ denote the probability of trajectory $o$ under the original policy. If $B_t\to\infty$ at every layer,
    then the distribution of sampled complete trajectories converges to
    \begin{equation}
        \label{eq:induced_distribution}
        \mu_\theta(o)
        \propto
        \pi_\theta(o)\exp\bigl(\gamma V(o)\bigr).
    \end{equation}
\end{theorem}

This result shows that our rollout mechanism does not sample trajectories directly from the original policy distribution $\pi_\theta$ (detailed proofs are presented in Appendix \ref{proof:1}). Instead, it induces a reweighted distribution that favors trajectories with larger informativeness values. In other words, informativeness acts as an intrinsic bias on the rollout process, encouraging the policy to allocate more computation to trajectories that are expected to be more informative during search. The following analysis further justifies such an intrinsic bias. In particular, we show that the induced distribution $\mu_\theta$ is in fact the optimizer of a variational problem.

\begin{theorem}[Variational characterization of the induced trajectory distribution]
    \label{theorem:2}
    Let $\pi_\theta$ denote the original policy-induced trajectory distribution, and let $V(o)$ denote the cumulative informativeness value of trajectory $o$. Consider the following variational objective over trajectory distributions $\mu$:
    \begin{equation}
        \label{eq:u_variational_objective}
        \max_{\mu}\;
        \mathbb{E}_{o\sim \mu}\!\left[V(o)\right]
        -\frac{1}{\gamma}\,\mathbb{D}_{\mathrm{KL}}\!\left(\mu \,\|\, \pi_\theta\right).
    \end{equation}
    Then the unique optimizer of \eqref{eq:u_variational_objective} is the induced trajectory $\mu_\theta(o)\propto \pi_\theta(o)\exp\bigl(\gamma\,V(o)\bigr).$
\end{theorem}

Theorem \ref{theorem:2} (see Appendix \ref{proof:2}) further provides a practical bridge to the GRPO objective in \eqref{eq:grpo-regularized}.
According to our preliminary observation in Fig.~\ref{fig:pre}, the trajectory reward $R(o)$ is strongly positively correlated with the cumulative value $V(o)$.  Therefore, as long as policy optimization is performed in a regime where the updated policy remains at a comparable distance from the reference policy $\pi_{\mathrm{ref}}$, the induced distribution $\mu_\theta$ can be viewed as a reward-aligned teacher that provides a useful policy improvement direction for the reward-based objective. Accordingly, we obtain a clear optimization target: \textbf{the distribution $\mu_{\theta}$ serves as a teacher distribution, specifying a policy improvement direction for the current policy $\pi_\theta$}.
Based on this view, we use the induced distribution as an informative sampling distribution for policy optimization.
At each training iteration, we freeze the current policy as $\pi_{\theta_{\mathrm{old}}}$ and construct the teacher sampler induced by the information gain-based rollout process in Section~\ref{sec:rollout}.
This sampler approximately draws trajectories from $\mu_{\theta_{\mathrm{old}}}$ under finite rollout budgets, thereby biasing the collected training samples toward trajectories with more informative intermediate states.
Given the sampled trajectories for each question, we then apply a group-based policy optimization objective to update $\pi_\theta$, where advantages provide the optimization signal and a KL penalty to $\pi_{\mathrm{ref}}$ is used for stable training.

\paragraph{Surrogate Objective.}
We now turn to the optimization objective induced by our information gain-based rollout.
For each question $q$, let $\{o_i\}_{i=1}^{G}$ denote the set of all sampled root-to-leaf trajectories, where $G$ is the number of leaf nodes and each path $o_i$ is associated with a reward $R(o_i)$. We define the following IGRPO objective:
\begin{equation}
    \begin{aligned}
         & \mathcal{J}_{\mathrm{IGRPO}}(\theta)
        =\mathbb{E}_{\substack{(q,a)\sim\mathcal{D},\\\{o_i\}\sim\mu_{\theta_{\mathrm{old}}}(\cdot\mid q)}}
        \Bigg[\\
            &\frac{1}{G}\sum_{i=1}^{G}\frac{1}{|o_i|}\sum_{t=1}^{T_i}\sum_{k=1}^{|\tau_{i,t}|}
            \min\Bigg(r_{i,t,k}\hat A_i,\,\\
            &\mathrm{clip}\Big(r_{i,t,k},1-\epsilon,1+\epsilon\Big)\hat A_i
            \Bigg)
            -\beta\mathbb{D}_{\mathrm{KL}}(\pi_\theta\|\pi_{\mathrm{ref}})
            \Bigg],
    \end{aligned}
    \label{eq:igrpo-surrogate}
\end{equation}
where $r_{i,t,k}=\frac{\pi_\theta(\tau_{i,t,k}\mid h_{i,t-1},\tau_{i,t,<k})}
    {\pi_{\theta_{\mathrm{old}}}(\tau_{i,t,k}\mid h_{i,t-1}, \tau_{i,t,<k})}$ is the importance ratio, $\hat A_i=
    \frac{R(o_i)-\mathrm{mean}\left(R(o_1),\dots,R(o_G)\right)}{\mathrm{std}\left(R(o_1),\dots,R(o_G)\right)}$ is the normalized group relative advantage of tree path $o_i$ and $\epsilon$ is the clipping parameter. $\mu_{\theta_{\mathrm{old}}}$ represents the information gain-based sampling process in Section \ref{sec:rollout}. For paths that terminate due to the absence of expandable nodes, we set their rewards according to the cumulative information gain along the path. For the node-level informativeness scoring function, we instantiate $\mathrm{val}(\cdot)$ as the arithmetic mean of the cumulative information gain and the instantaneous information gain:
\begin{equation}
    \label{eq:val}
    \mathrm{val}(h)
    =
    \frac{s_\theta(h; a) + \mathrm{IG}(h)}{2}.
\end{equation}
This captures both the current promise of the node and the marginal progress made by the latest transition, encouraging the rollout selector to prioritize nodes that are both answerable and still improving. The complete IGRPO workflow is in Appendix \ref{sec:algorithm}.

\section{Experiments}
\begin{table*}[t]
    \centering
    \fontsize{10pt}{12pt}\selectfont
    \setlength{\tabcolsep}{1pt}
    \begin{tabular*}{\textwidth}{@{\extracolsep{\fill}}l@{\hspace{4pt}}l*{16}{c}@{}}
        \toprule
        \multirow{3}{*}{Type} & \multirow{3}{*}{Method}
        & \multicolumn{6}{c}{\textbf{Single-Hop QA}}
        & \multicolumn{8}{c}{\textbf{Multi-Hop QA}}
        & \multicolumn{2}{c}{\multirow{2}{*}{Avg.}} \\
        \cmidrule(lr){3-8} \cmidrule(lr){9-16}
        & & \multicolumn{2}{c}{NQ$^{\dagger}$}
        & \multicolumn{2}{c}{TriviaQA$^{\star}$}
        & \multicolumn{2}{c}{PopQA$^{\star}$}
        & \multicolumn{2}{c}{HotpotQA$^{\dagger}$}
        & \multicolumn{2}{c}{2Wiki$^{\star}$}
        & \multicolumn{2}{c}{MuSiQue$^{\star}$}
        & \multicolumn{2}{c}{Bamboogle$^{\star}$}
        & \multicolumn{2}{c}{} \\
        \cmidrule(lr){3-4} \cmidrule(lr){5-6} \cmidrule(lr){7-8} \cmidrule(lr){9-10} \cmidrule(lr){11-12} \cmidrule(lr){13-14} \cmidrule(lr){15-16} \cmidrule(lr){17-18}
        & & \textit{3B} & \textit{7B} & \textit{3B} & \textit{7B} & \textit{3B} & \textit{7B} & \textit{3B} & \textit{7B} & \textit{3B} & \textit{7B} & \textit{3B} & \textit{7B} & \textit{3B} & \textit{7B} & \textit{3B} & \textit{7B} \\
        \midrule
        Vanilla & R1-Instruct & 27.0 & 21.0 & 53.7 & 44.9 & 19.9 & 17.1 & 23.7 & 20.8 & 29.2 & 27.5 & 7.2 & 6.0 & 29.6 & 19.2 & 27.2 & 22.4 \\
        Chain & Search-R1 & 34.1 & 39.3 & 54.5 & 61.0 & 37.8 & 39.7 & 32.4 & 37.0 & 31.9 & 40.1 & 10.3 & 14.6 & 26.4 & 36.8 & 32.5 & 38.4 \\
        Chain & ZeroSearch & 41.4 & 43.6 & 57.4 & 61.8 & \underline{44.8} & \textbf{51.5} & 27.4 & 34.6 & 30.0 & 35.2 & 9.8 & 18.4 & 11.2 & 28.0 & 31.7 & 39.0 \\
        Chain & StepSearch & -- & -- & -- & -- & -- & -- & 34.5 & 38.6 & 32.0 & 36.6 & \textbf{17.4} & \textbf{22.6} & \textbf{34.4} & 40.0 & -- & -- \\
        Chain & GiGPO & 42.0 & \underline{46.4} & 59.5 & \textbf{64.7} & 42.4 & 46.1 & \underline{36.9} & 41.6 & \underline{37.0} & \underline{43.6} & 12.6 & 18.9 & 28.8 & 38.4 & \underline{37.0} & \underline{42.8} \\
        Chain & IGPO & 40.2 & 44.0 & 58.1 & 60.1 & 43.2 & 45.4 & 35.7 & \underline{42.9} & 34.9 & 42.5 & 12.3 & 16.8 & 29.6 & 40.0 & 36.3 & 41.7 \\
        Tree & Tree-GRPO & \underline{44.5} & 45.6 & \underline{60.4} & \underline{62.5} & 43.7 & 42.8 & 31.1 & 39.9 & 31.7 & 35.6 & 11.9 & 17.5 & 28.0 & \underline{41.6} & 35.9 & 40.8 \\
        Tree & \textbf{IGRPO} & \textbf{45.6} & \textbf{48.0} & \textbf{60.9} & \textbf{64.7} & \textbf{47.5} & \underline{47.0} & \textbf{38.9} & \textbf{44.6} & \textbf{40.2} & \textbf{43.8} & \underline{14.1} & \underline{20.2} & \underline{32.0} & \textbf{44.0} & \textbf{39.9} & \textbf{44.6} \\
        \bottomrule
    \end{tabular*}
    \caption{Performance on seven search-augmented QA tasks (\%) using Qwen2.5-3B/7B-Instruct. The best results are highlighted in \textbf{bold}, and the runner-up results are \underline{underlined}. $\dagger$ and $\star$ indicate in-domain and out-of-domain datasets, respectively. Avg. denotes the arithmetic mean of the EM scores across all datasets.}
    \label{tab:main_results}
\end{table*}

\subsection{Experiment Setup}
\paragraph{Benchmarks \& Metrics.}
We evaluate IGRPO on seven search-augmented QA benchmarks covering single-hop and multi-hop question answering tasks. Single-hop benchmarks include TriviaQA~\cite{joshi2017triviaqa}, NQ~\cite{kwiatkowski2019natural} and PopQA~\cite{mallen2023not}, while the more challenging multi-hop setting includes HotpotQA~\cite{yang2018hotpotqa}, 2Wiki~\cite{ho2020constructing}, MusiQue~\cite{trivedi2022musique}, and Bamboogle~\cite{press2023measuring}, which require the model to perform multi-step reasoning by integrating information across multiple pieces of evidence. We use exact match (EM) as the evaluation metric, which measures whether the predicted answer matches the reference answer after normalization.

\paragraph{Baselines.} We compare IGRPO with representative baselines. Our evaluation includes R1-Instruct as the vanilla instruction-tuned backbone, as well as Search-R1~\cite{jin2025search} and ZeroSearch~\cite{sun2025zerosearch} as two chain-based search RL methods with trajectory-level optimization. We further consider representative chain-based methods with finer-grained step-level optimization, including StepSearch~\cite{wang2025stepsearch}, GiGPO~\cite{feng2025group}, and IGPO~\cite{wang2025information}. In addition, we include the tree-based baseline Tree-GRPO~\cite{ji2025tree}, which extends group-based policy optimization to tree-structured rollouts. These baselines span instruction tuning and chain- and tree-based RL.
\paragraph{Training details.} All baselines share the same experimental configuration, using Qwen2.5-3B/7B-Instruct~\cite{yang2024qwen2} as the base models. For comparable search budgets, we set the rollout group size to $N=5$ for chain-based baselines and the per-stage budget to $B_i=5$ for IGRPO. Following Search-R1, we use E5~\cite{wang2022text} as the retriever, and set the maximum number of interaction turns to 4. All models are trained for 200 iterations, with further details in Appendix~\ref{sec:training_details}.

\subsection{Overall Performance}

\paragraph{Main results.} Table~\ref{tab:main_results} presents the main results on seven search-augmented QA benchmarks.
Overall, IGRPO achieves the highest average EM at both backbone scales, reaching 39.9\% on 3B and 44.6\% on 7B. These results show that our information gain-based rollout can consistently improve agent training across both single-hop and multi-hop tasks.
Compared with strong chain-based methods, IGRPO provides additional gains by going beyond trajectory-level or turn-level credit assignment and explicitly reallocating rollout computation.

The average improvement over GiGPO is larger on the 3B backbone (2.9\%) than on the 7B backbone (1.8\%). This suggests that smaller models benefit more from rollout guidance, since their initial search policies are less reliable and are more likely to spend computation on low-value branches. By evaluating intermediate nodes, IGRPO helps the model focus on more promising states, thereby providing stronger supervision.

\paragraph{Comparison under varying rollout budgets.} Notably, IGRPO outperforms both IGPO and Tree-GRPO, suggesting that sampling more informative trajectories is more effective than using information gain to shape rewards for conventionally sampled trajectories or expanding trees randomly.
The advantage of IGRPO comes from its clearer optimization target: the information gain-based rollout induces a limiting teacher distribution, which favors informative training trajectories. The average EM results in Table~\ref{tab:rollout_budget} further support this benefit: IGPO with an increased budget ($N=6$) still trails IGRPO-low, while IGRPO with a reduced budget ($B_i=4$) outperforms default IGPO ($N=5$). These results suggest that reshaping the sampling distribution is more effective than merely drawing more trajectories from the original policy.

\begin{table}[t]
    \centering
    \fontsize{10pt}{12pt}\selectfont
    \setlength{\tabcolsep}{1pt}
    \begin{tabular*}{\columnwidth}{@{\extracolsep{\fill}}l*{5}{c}@{}}
        \toprule
        Method & Hotpot. & 2Wiki & MuSi. & Bamb. & Avg. \\
        \midrule
        IGPO ($N=5$)    & 35.7 & 34.9 & 12.3 & 29.6 & 28.1 \\
        IGRPO ($B_i=4$) & 36.9 & 38.5 & 12.1 & 31.2 & 29.7 \\
        IGPO ($N=6$)    & 38.1 & 34.5 & \textbf{14.6} & \textbf{34.4} & 30.4 \\
        IGRPO-low      & \textbf{39.4} & \underline{38.7} & \underline{14.4} & 31.2 & \underline{30.9} \\
        IGRPO ($B_i=5$) & \underline{38.9} & \textbf{40.2} & 14.1 & \underline{32.0} & \textbf{31.3} \\
        \bottomrule
    \end{tabular*}
    \caption{EM (\%) on multi-hop QA under different rollout budgets, with all models trained on the 3B backbone. IGRPO-low denotes the variant with a per-stage budget set to the current active set size.}
    \label{tab:rollout_budget}
\end{table}

\begin{figure}[t]
    \centering
    \includegraphics[width=\linewidth]{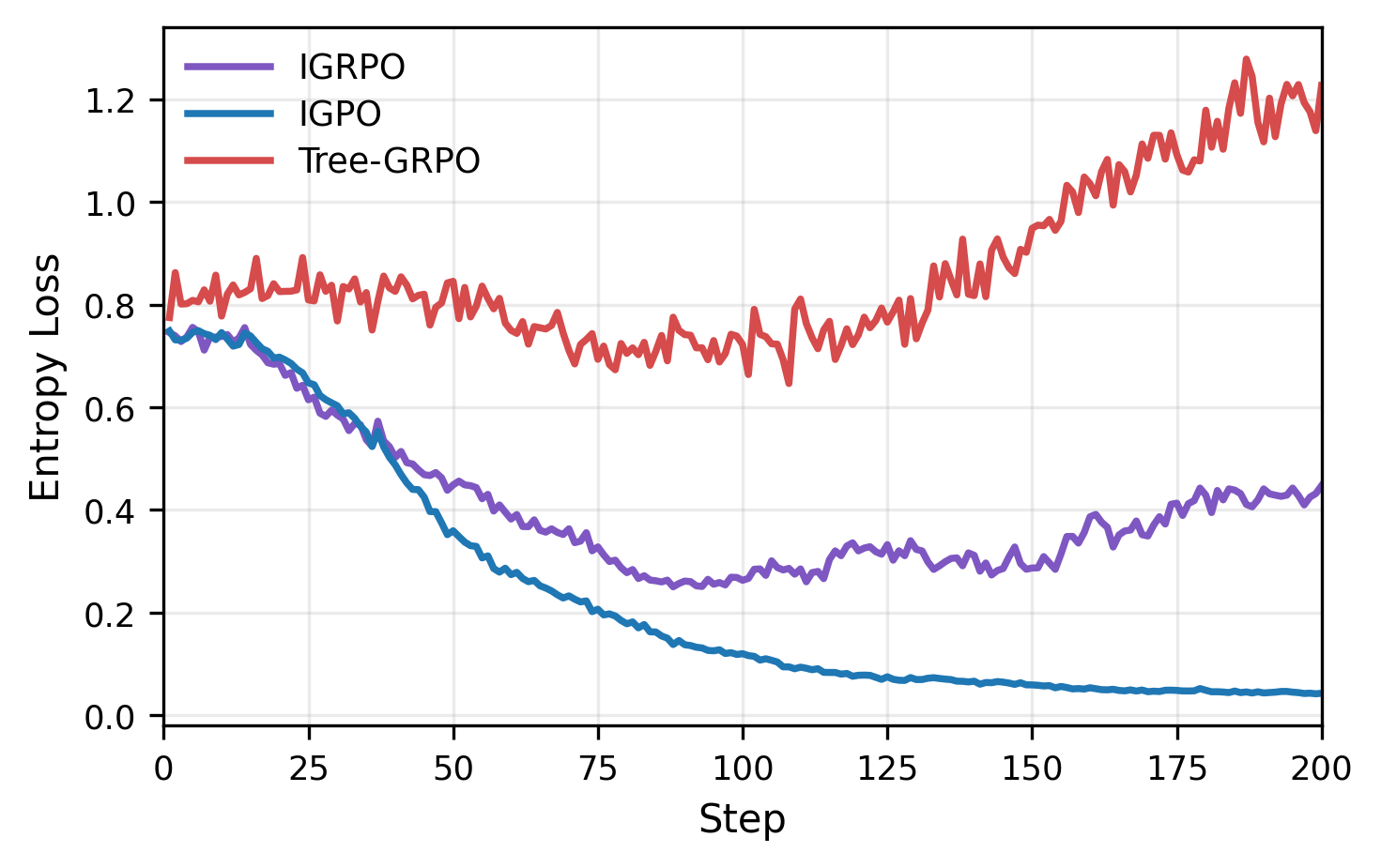}
    \caption{Entropy during training on 3B backbone.}
    \label{fig:main_entropy}
\end{figure}

\paragraph{Entropy dynamics.} We further examine the entropy dynamics of different strategies in Figure~\ref{fig:main_entropy}. IGPO rapidly reduces entropy, suggesting premature concentration on a narrow range of behaviors. In contrast, Tree-GRPO maintains high entropy, suggesting that random tree expansion increases exploration but also amplifies the model's uncertainty,  making it harder for the model to converge toward stable search behaviors. IGRPO's initial decrease suggests improved search capability, while the subsequent slight increase reflects exploration of promising regions of the search space. Additional analysis of entropy dynamics is provided in Appendix~\ref{sec:entropy_analysis}, and case studies in Appendix~\ref{sec:case_study} further illustrate IGRPO's search capabilities.

\subsection{Ablation Study}
\begin{figure}[t]
    \centering
    \includegraphics[width=\linewidth]{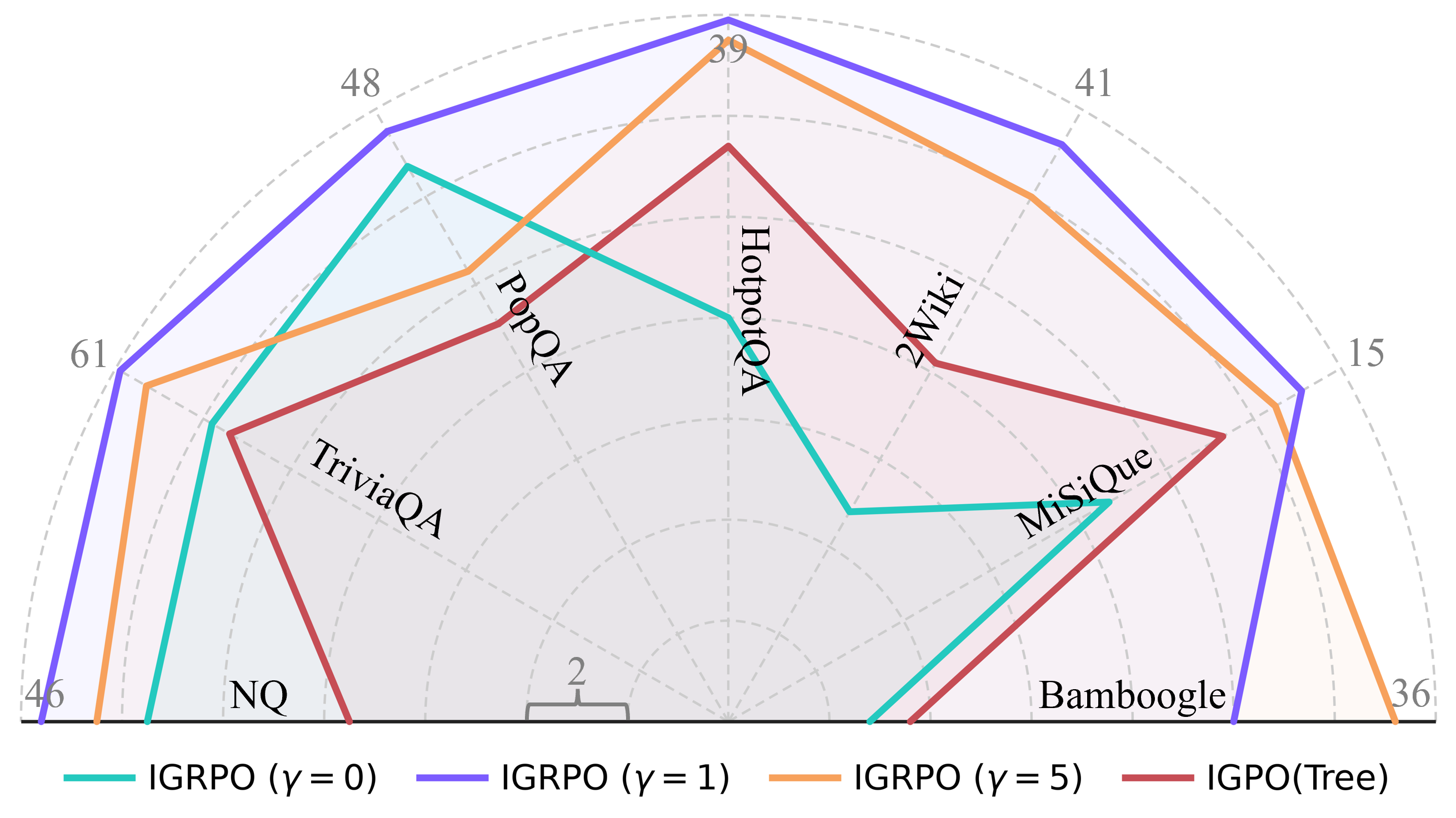}
    \caption{Ablation results for IGRPO ($\gamma\in\{0,1,5\}$) and IGPO with random tree expansion.}
    \label{fig:ablation_gamma}
\end{figure}

The temperature $\gamma$ controls the trade-off between favoring informative trajectories and preserving exploration diversity.
Figure~\ref{fig:ablation_gamma} compares three $\gamma$ values and IGPO(Tree) on the 3B backbone. With $\gamma=0$, uniform node selection resembles Tree-GRPO's random expansion and yields 35.5\% average EM across seven datasets. The $\gamma=5$ variant places greater emphasis on highly informative trajectories but performs worse than the moderate $\gamma=1$ variant on most datasets, suggesting that the latter offers a favorable balance between informativeness and exploration. With $\gamma=5$, concentrating expansion on high-scoring nodes may limit opportunities to explore alternative trajectories.
IGPO(Tree), which combines random tree expansion with information gain-based credit assignment, achieves 35.8\% average EM. IGRPO ($\gamma=1$) outperforms IGPO(Tree) on all seven datasets, indicating benefits beyond tree-structured rollouts alone. This comparison supports using information gain to guide selection rather than solely for credit assignment, consistent with our focus on \textbf{how training trajectories are generated}. Additional ablation results on the criterion $\mathrm{val}(\cdot)$ and reward $R(\cdot)$ design details are provided in Appendix~\ref{sec:additional_ablation_study}.

\subsection{Computational Analysis}
\label{sec:efficiency}

\begin{figure}[t]
    \centering
    \includegraphics[width=\linewidth]{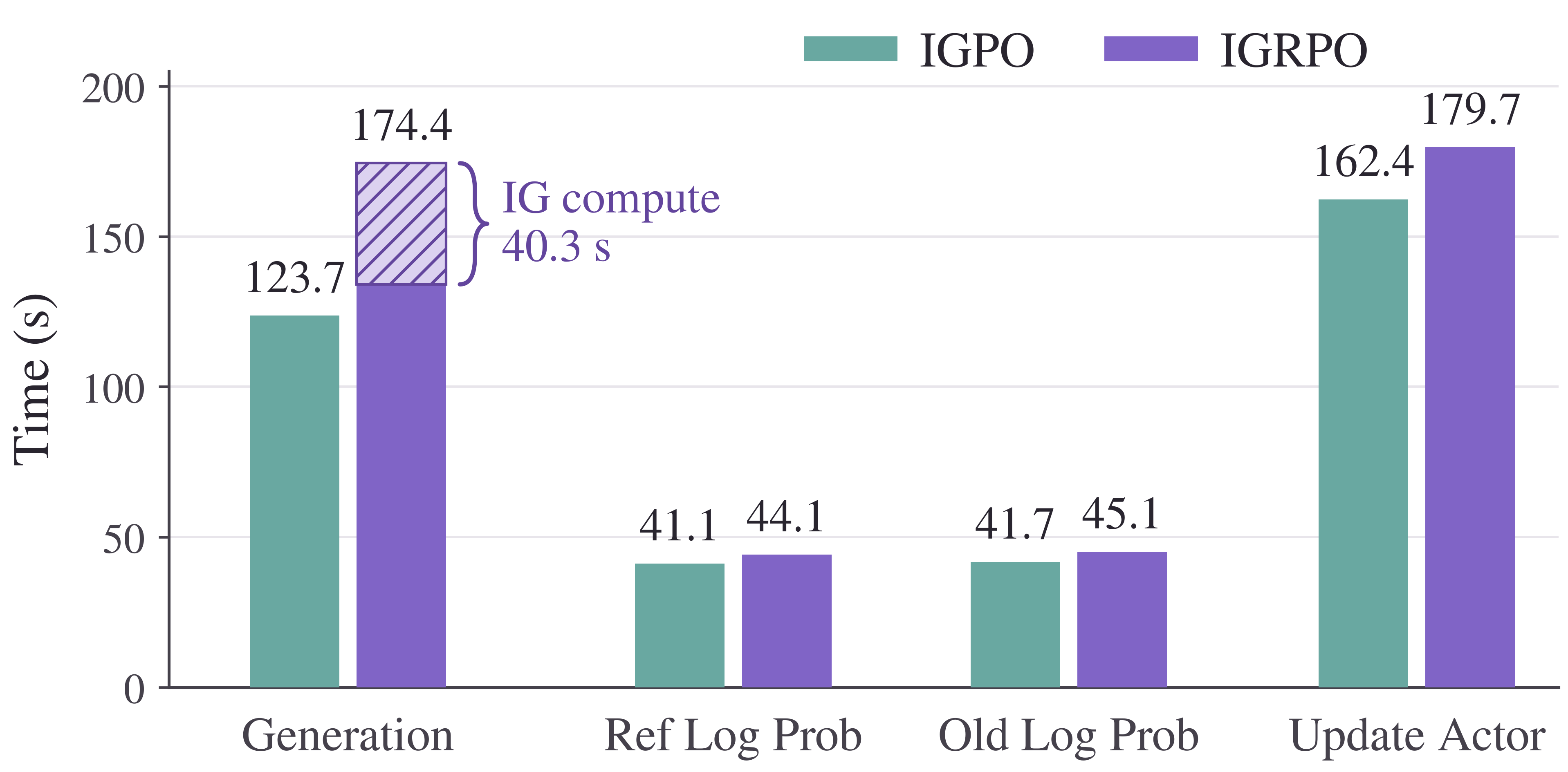}
    \caption{Average time breakdown of IGPO and IGRPO across major training components within one step.}
    \label{fig:time_breakdown}
\end{figure}

\paragraph{Training stage.} To assess the comparability of training budgets, we analyze the time spent on each major training component. Relative to standard search rollouts, IGRPO adds one batched information gain computation after each search step. As shown in Figure~\ref{fig:time_breakdown}, IGRPO requires 443.3\,s per training step, compared with 368.9\,s for IGPO, whose training time is comparable to that of other baselines since it incorporates IG computation into other part. IGRPO incurs a 20.2\% overhead, with over half attributable to the 40.3\,s spent on IG computation. This computation guides the generation of more beneficial training trajectories, yielding performance gains that justify the additional cost, as demonstrated by the preceding experiments.

\paragraph{Inference stage.} By prioritizing informative trajectories at training stage, IG-guided selection encourages more reliable search and more effective synthesis of retrieved evidence. As shown in Table~\ref{tab:tool_call}, IGRPO generally requires fewer tool calls than IGPO and Tree-GRPO at inference. Together with the preceding accuracy gains, this reduction suggests more efficient use of retrieval.

\begin{table}[t]
    \centering
    \fontsize{10pt}{12pt}\selectfont
    \setlength{\tabcolsep}{1pt}
    \begin{tabular*}{\columnwidth}{@{\extracolsep{\fill}}l*{5}{c}@{}}
        \toprule
        Method & Trivia. & Pop. & Hotpot. & 2Wiki & MuSi. \\
        \midrule
        IGPO      & 1.36 & 1.47 & 1.68 & 1.93 & 2.39 \\
        Tree-GRPO & 1.24 & 1.17 & 1.71 & 1.89 & 2.43 \\
        IGRPO     & 1.15 & 1.22 & 1.50 & 1.86 & 1.94 \\
        \bottomrule
    \end{tabular*}
    \caption{Average number of tool calls during inference.}
    \label{tab:tool_call}
\end{table}

\section{Conclusion}
We propose IGRPO, a tree-structured rollout framework for search agents that uses information gain to adaptively allocate computation to informative intermediate states. By shaping how training trajectories are generated, IGRPO favors trajectories with higher cumulative informativeness. Our theoretical analysis characterizes the limiting trajectory distribution as a reweighted policy distribution, motivating its use as a teacher for policy optimization. Experiments on seven search-augmented QA benchmarks with 3B and 7B models demonstrate improved accuracy under comparable rollout budgets. Ablations support using information gain for rollout selection rather than solely for credit assignment, while generally fewer tool calls at inference suggest more efficient retrieval.

\section*{Limitations}
Our current information gain measure relies on a reference answer, limiting its direct applicability to open-ended tasks with diverse valid responses or no definitive reference answer. A smaller scoring model could provide alternative node assessments for the same allocation mechanism, which warrants future investigation.
Resource constraints limit our experiments to modest rollout budgets, whereas the theory assumes an asymptotic regime. Greater computational resources would enable larger budgets and potentially a more stable approximation of the induced trajectory distribution.

\bibliography{ref}

@article{shao2024deepseekmath,
  title   = {Deepseekmath: Pushing the limits of mathematical reasoning in open language models},
  author  = {Shao, Zhihong and Wang, Peiyi and Zhu, Qihao and Xu, Runxin and Song, Junxiao and Bi, Xiao and Zhang, Haowei and Zhang, Mingchuan and Li, YK and others},
  journal = {arXiv preprint arXiv:2402.03300},
  year    = {2024}
}

@article{schulman2017proximal,
  title   = {Proximal policy optimization algorithms},
  author  = {Schulman, John and Wolski, Filip and Dhariwal, Prafulla and Radford, Alec and Klimov, Oleg},
  journal = {arXiv preprint arXiv:1707.06347},
  year    = {2017}
}

@inproceedings{jin2025search,
  title     = {Search-R1: Training {LLM}s to Reason and Leverage Search Engines with Reinforcement Learning},
  author    = {Bowen Jin and Hansi Zeng and Zhenrui Yue and Jinsung Yoon and Sercan O Arik and Dong Wang and Hamed Zamani and Jiawei Han},
  booktitle = {Second Conference on Language Modeling},
  year      = {2025},
  url       = {https://openreview.net/forum?id=Rwhi91ideu}
}

@article{sun2025zerosearch,
  title   = {Zerosearch: Incentivize the search capability of llms without searching},
  author  = {Sun, Hao and Qiao, Zile and Guo, Jiayan and Fan, Xuanbo and Hou, Yingyan and Jiang, Yong and Xie, Pengjun and Zhang, Yan and Huang, Fei and Zhou, Jingren},
  journal = {arXiv preprint arXiv:2505.04588},
  year    = {2025}
}

@inproceedings{wang2025stepsearch,
  title     = {{S}tep{S}earch: Igniting {LLM}s Search Ability via Step-Wise Proximal Policy Optimization},
  author    = {Zheng, Xuhui  and
               An, Kang  and
               Wang, Ziliang  and
               Wang, Yuhang  and
               Wu, Yichao},
  editor    = {Christodoulopoulos, Christos  and
               Chakraborty, Tanmoy  and
               Rose, Carolyn  and
               Peng, Violet},
  booktitle = {Proceedings of the 2025 Conference on Empirical Methods in Natural Language Processing},
  month     = nov,
  year      = {2025},
  address   = {Suzhou, China},
  publisher = {Association for Computational Linguistics},
  url       = {https://aclanthology.org/2025.emnlp-main.1106/},
  doi       = {10.18653/v1/2025.emnlp-main.1106},
  pages     = {21805--21830},
  isbn      = {979-8-89176-332-6}
}

@inproceedings{feng2025group,
  author    = {Feng, Lang and Xue, Zhenghai and Liu, Tingcong and An, Bo},
  booktitle = {Advances in Neural Information Processing Systems},
  doi       = {10.52202/085713-1544},
  editor    = {D. Belgrave and C. Zhang and H. Lin and R. Pascanu and P. Koniusz and M. Ghassemi and N. Chen},
  pages     = {46375--46408},
  publisher = {Curran Associates, Inc.},
  title     = {Group-in-Group Policy Optimization for LLM Agent Training},
  url       = {https://proceedings.neurips.cc/paper_files/paper/2025/file/420c9f777c0b4f78d515e53cf74d58b2-Paper-Conference.pdf},
  volume    = {38, Main Conference},
  year      = {2025}
}

@inproceedings{wang2025information,
  title     = {Information Gain-based Policy Optimization: A Simple and Effective Approach for Multi-Turn Search Agents},
  author    = {Guoqing Wang and Sunhao Dai and Guangze Ye and Zeyu Gan and Wei Yao and Yong Deng and Xiaofeng Wu and Zhenzhe Ying},
  booktitle = {The Fourteenth International Conference on Learning Representations},
  year      = {2026},
  url       = {https://openreview.net/forum?id=qkWP6phrvZ}
}

@inproceedings{dong2025agentic,
  title     = {Agentic Reinforced Policy Optimization},
  author    = {Guanting Dong and Hangyu Mao and Kai Ma and Licheng Bao and Yifei Chen and Zhongyuan Wang and Zhongxia Chen and Jiazhen Du and Huiyang Wang and Fuzheng Zhang and Guorui Zhou and Yutao Zhu and Ji-Rong Wen and Zhicheng Dou},
  booktitle = {The Fourteenth International Conference on Learning Representations},
  year      = {2026},
  url       = {https://openreview.net/forum?id=TX4k7BF6aO}
}

@inproceedings{ji2025tree,
  title     = {Tree Search for {LLM} Agent Reinforcement Learning},
  author    = {Yuxiang Ji and Ziyu Ma and Yong Wang and Guanhua Chen and Xiangxiang Chu and Liaoni Wu},
  booktitle = {The Fourteenth International Conference on Learning Representations},
  year      = {2026},
  url       = {https://openreview.net/forum?id=ZpQwAFhU13}
}

@article{kwiatkowski2019natural,
  title     = {Natural questions: a benchmark for question answering research},
  author    = {Kwiatkowski, Tom and Palomaki, Jennimaria and Redfield, Olivia and Collins, Michael and Parikh, Ankur and Alberti, Chris and Epstein, Danielle and Polosukhin, Illia and Devlin, Jacob and Lee, Kenton and others},
  journal   = {Transactions of the Association for Computational Linguistics},
  volume    = {7},
  pages     = {453--466},
  year      = {2019},
  publisher = {MIT Press One Rogers Street, Cambridge, MA 02142-1209, USA journals-info~…}
}

@inproceedings{joshi2017triviaqa,
  title     = {Triviaqa: A large scale distantly supervised challenge dataset for reading comprehension},
  author    = {Joshi, Mandar and Choi, Eunsol and Weld, Daniel S and Zettlemoyer, Luke},
  booktitle = {Proceedings of the 55th Annual Meeting of the Association for Computational Linguistics (Volume 1: Long Papers)},
  pages     = {1601--1611},
  year      = {2017}
}

@inproceedings{mallen2023not,
  title     = {When not to trust language models: Investigating effectiveness of parametric and non-parametric memories},
  author    = {Mallen, Alex and Asai, Akari and Zhong, Victor and Das, Rajarshi and Khashabi, Daniel and Hajishirzi, Hannaneh},
  booktitle = {Proceedings of the 61st annual meeting of the association for computational linguistics (volume 1: Long papers)},
  pages     = {9802--9822},
  year      = {2023}
}

@inproceedings{yang2018hotpotqa,
  title     = {HotpotQA: A dataset for diverse, explainable multi-hop question answering},
  author    = {Yang, Zhilin and Qi, Peng and Zhang, Saizheng and Bengio, Yoshua and Cohen, William and Salakhutdinov, Ruslan and Manning, Christopher D},
  booktitle = {Proceedings of the 2018 conference on empirical methods in natural language processing},
  pages     = {2369--2380},
  year      = {2018}
}

@inproceedings{ho2020constructing,
  title     = {Constructing a multi-hop qa dataset for comprehensive evaluation of reasoning steps},
  author    = {Ho, Xanh and Nguyen, Anh-Khoa Duong and Sugawara, Saku and Aizawa, Akiko},
  booktitle = {Proceedings of the 28th International Conference on Computational Linguistics},
  pages     = {6609--6625},
  year      = {2020}
}

@article{trivedi2022musique,
  title     = {MuSiQue: Multihop Questions via Single-hop Question Composition},
  author    = {Trivedi, Harsh and Balasubramanian, Niranjan and Khot, Tushar and Sabharwal, Ashish},
  journal   = {Transactions of the Association for Computational Linguistics},
  volume    = {10},
  pages     = {539--554},
  year      = {2022},
  publisher = {MIT Press One Broadway, 12th Floor, Cambridge, Massachusetts 02142, USA~…}
}

@inproceedings{press2023measuring,
  title     = {Measuring and narrowing the compositionality gap in language models},
  author    = {Press, Ofir and Zhang, Muru and Min, Sewon and Schmidt, Ludwig and Smith, Noah A and Lewis, Mike},
  booktitle = {Findings of the Association for Computational Linguistics: EMNLP 2023},
  pages     = {5687--5711},
  year      = {2023}
}

@misc{yang2024qwen2,
  title         = {Qwen2.5 Technical Report},
  author        = {Qwen and : and An Yang and Baosong Yang and Beichen Zhang and Binyuan Hui and Bo Zheng and Bowen Yu and Chengyuan Li and Dayiheng Liu and Fei Huang and Haoran Wei and Huan Lin and Jian Yang and Jianhong Tu and Jianwei Zhang and Jianxin Yang and Jiaxi Yang and Jingren Zhou and Junyang Lin and Kai Dang and Keming Lu and Keqin Bao and Kexin Yang and Le Yu and Mei Li and Mingfeng Xue and Pei Zhang and Qin Zhu and Rui Men and Runji Lin and Tianhao Li and Tianyi Tang and Tingyu Xia and Xingzhang Ren and Xuancheng Ren and Yang Fan and Yang Su and Yichang Zhang and Yu Wan and Yuqiong Liu and Zeyu Cui and Zhenru Zhang and Zihan Qiu},
  year          = {2025},
  eprint        = {2412.15115},
  archiveprefix = {arXiv},
  primaryclass  = {cs.CL},
  url           = {https://arxiv.org/abs/2412.15115}
}

@article{wang2022text,
  title   = {Text embeddings by weakly-supervised contrastive pre-training},
  author  = {Wang, Liang and Yang, Nan and Huang, Xiaolong and Jiao, Binxing and Yang, Linjun and Jiang, Daxin and Majumder, Rangan and Wei, Furu},
  journal = {arXiv preprint arXiv:2212.03533},
  year    = {2022}
}

@article{donsker1975asymptotic,
  title     = {Asymptotic evaluation of certain Markov process expectations for large time, I},
  author    = {Donsker, Monroe D and Varadhan, SR Srinivasa},
  journal   = {Communications on pure and applied mathematics},
  volume    = {28},
  number    = {1},
  pages     = {1--47},
  year      = {1975},
  publisher = {Wiley Online Library}
}

@article{guo2025deepseek,
  title   = {Deepseek-r1: Incentivizing reasoning capability in llms via reinforcement learning},
  author  = {Guo, Daya and Yang, Dejian and Zhang, Haowei and Song, Junxiao and Wang, Peiyi and Zhu, Qihao and Xu, Runxin and Zhang, Ruoyu and Ma, Shirong and Bi, Xiao and others},
  journal = {arXiv preprint arXiv:2501.12948},
  year    = {2025}
}

@article{ouyang2022training,
  title   = {Training language models to follow instructions with human feedback},
  author  = {Ouyang, Long and Wu, Jeffrey and Jiang, Xu and Almeida, Diogo and Wainwright, Carroll and Mishkin, Pamela and Zhang, Chong and Agarwal, Sandhini and Slama, Katarina and Ray, Alex and others},
  journal = {Advances in neural information processing systems},
  volume  = {35},
  pages   = {27730--27744},
  year    = {2022}
}

@inproceedings{ahmadian2024back,
  title     = {Back to basics: Revisiting REINFORCE-style optimization for learning from human feedback in LLMs},
  author    = {Ahmadian, Arash and Cremer, Chris and Gall{\'e}, Matthias and Fadaee, Marzieh and Kreutzer, Julia and Pietquin, Olivier and {\"U}st{\"u}n, Ahmet and Hooker, Sara},
  booktitle = {Proceedings of the 62nd Annual Meeting of the Association for Computational Linguistics (Volume 1: Long Papers)},
  pages     = {12248--12267},
  year      = {2024}
}

@inproceedings{yu2025dapo,
  author    = {Yu, Qiying and Zhang, Zheng and Zhu, Ruofei and Yuan, Yufeng and Zuo, Xiaochen and Yue, Yu and Dai, Weinan and Fan, Tiantian and Liu, Gaohong and liu, juncai and Liu, LingJun and Liu, Xin and Lin, Haibin and Lin, Zhiqi and Ma, Bole and Sheng, Guangming and Tong, Yuxuan and Zhang, Chi and Zhang, Mofan and Zhang, Ru and Zhang, Wang and Zhu, Hang and Zhu, Jinhua and Chen, Jiaze and Chen, Jiangjie and Wang, Chengyi and Yu, Hongli and Song, Yuxuan and Wei, Xiangpeng and Zhou, Hao and Liu, Jingjing and Ma, Wei-Ying and Zhang, Ya-Qin and Yan, Lin and Wu, Yonghui and Wang, Mingxuan},
  booktitle = {Advances in Neural Information Processing Systems},
  doi       = {10.52202/085713-3775},
  editor    = {D. Belgrave and C. Zhang and H. Lin and R. Pascanu and P. Koniusz and M. Ghassemi and N. Chen},
  pages     = {113222--113244},
  publisher = {Curran Associates, Inc.},
  title     = {DAPO: An Open-Source LLM Reinforcement Learning System at Scale},
  url       = {https://proceedings.neurips.cc/paper_files/paper/2025/file/a4277440d50f1f15d2cb4c14f7e0c0d2-Paper-Conference.pdf},
  volume    = {38, Main Conference},
  year      = {2025}
}

@article{zheng2025group,
  title   = {Group sequence policy optimization},
  author  = {Zheng, Chujie and Liu, Shixuan and Li, Mingze and Chen, Xiong-Hui and Yu, Bowen and Gao, Chang and Dang, Kai and Liu, Yuqiong and Men, Rui and Yang, An and others},
  journal = {arXiv preprint arXiv:2507.18071},
  year    = {2025}
}

@article{song2025r1,
  title   = {R1-searcher: Incentivizing the search capability in llms via reinforcement learning},
  author  = {Song, Huatong and Jiang, Jinhao and Min, Yingqian and Chen, Jie and Chen, Zhipeng and Zhao, Wayne Xin and Fang, Lei and Wen, Ji-Rong},
  journal = {arXiv preprint arXiv:2503.05592},
  year    = {2025}
}

@inproceedings{song2025smart,
  title     = {Smart-Searcher: Incentivizing the Dynamic Knowledge Acquisition of {LLM}s via Reinforcement Learning},
  author    = {Song, Huatong  and
               Jiang, Jinhao  and
               Tian, Wenqing  and
               Chen, Zhipeng  and
               Wu, Yuhuan  and
               Zhao, Jiahao  and
               Min, Yingqian  and
               Zhao, Wayne Xin  and
               Fang, Lei  and
               Wen, Ji-Rong},
  editor    = {Christodoulopoulos, Christos  and
               Chakraborty, Tanmoy  and
               Rose, Carolyn  and
               Peng, Violet},
  booktitle = {Findings of the Association for Computational Linguistics: EMNLP 2025},
  month     = nov,
  year      = {2025},
  address   = {Suzhou, China},
  publisher = {Association for Computational Linguistics},
  url       = {https://aclanthology.org/2025.findings-emnlp.731/},
  doi       = {10.18653/v1/2025.findings-emnlp.731},
  pages     = {13572--13586},
  isbn      = {979-8-89176-335-7}
}

@article{dong2025aepo,
  title   = {Agentic entropy-balanced policy optimization},
  author  = {Dong, Guanting and Bao, Licheng and Wang, Zhongyuan and Zhao, Kangzhi and Li, Xiaoxi and Jin, Jiajie and Yang, Jinghan and Mao, Hangyu and Zhang, Fuzheng and Gai, Kun and others},
  journal = {arXiv preprint arXiv:2510.14545},
  year    = {2025}
}

@article{brown2020language,
  title   = {Language models are few-shot learners},
  author  = {Brown, Tom and Mann, Benjamin and Ryder, Nick and Subbiah, Melanie and Kaplan, Jared D and Dhariwal, Prafulla and Neelakantan, Arvind and Shyam, Pranav and Sastry, Girish and Askell, Amanda and others},
  journal = {Advances in neural information processing systems},
  volume  = {33},
  pages   = {1877--1901},
  year    = {2020}
}

@article{chowdhery2023palm,
  title   = {Palm: Scaling language modeling with pathways},
  author  = {Chowdhery, Aakanksha and Narang, Sharan and Devlin, Jacob and Bosma, Maarten and Mishra, Gaurav and Roberts, Adam and Barham, Paul and Chung, Hyung Won and Sutton, Charles and Gehrmann, Sebastian and others},
  journal = {Journal of machine learning research},
  volume  = {24},
  number  = {240},
  pages   = {1--113},
  year    = {2023}
}

@misc{touvron2023llama,
  title         = {LLaMA: Open and Efficient Foundation Language Models},
  author        = {Hugo Touvron and Thibaut Lavril and Gautier Izacard and Xavier Martinet and Marie-Anne Lachaux and Timothée Lacroix and Baptiste Rozière and Naman Goyal and Eric Hambro and Faisal Azhar and Aurelien Rodriguez and Armand Joulin and Edouard Grave and Guillaume Lample},
  year          = {2023},
  eprint        = {2302.13971},
  archiveprefix = {arXiv},
  primaryclass  = {cs.CL},
  url           = {https://arxiv.org/abs/2302.13971}
}

@inproceedings{yao2022react,
  title     = {ReAct: Synergizing Reasoning and Acting in Language Models},
  author    = {Shunyu Yao and Jeffrey Zhao and Dian Yu and Nan Du and Izhak Shafran and Karthik R Narasimhan and Yuan Cao},
  booktitle = {The Eleventh International Conference on Learning Representations },
  year      = {2023},
  url       = {https://openreview.net/forum?id=WE_vluYUL-X}
}

@article{schick2023toolformer,
  title   = {Toolformer: Language models can teach themselves to use tools},
  author  = {Schick, Timo and Dwivedi-Yu, Jane and Dess{\`\i}, Roberto and Raileanu, Roberta and Lomeli, Maria and Hambro, Eric and Zettlemoyer, Luke and Cancedda, Nicola and Scialom, Thomas},
  journal = {Advances in neural information processing systems},
  volume  = {36},
  pages   = {68539--68551},
  year    = {2023}
}

@inproceedings{qian2025toolrl,
  author    = {Qian, Cheng and Acikgoz, Emre Can and He, Qi and WANG, Hongru and Chen, Xiusi and Hakkani-Tur, Dilek and Tur, Gokhan and Ji, Heng},
  booktitle = {Advances in Neural Information Processing Systems},
  doi       = {10.52202/085713-3524},
  editor    = {D. Belgrave and C. Zhang and H. Lin and R. Pascanu and P. Koniusz and M. Ghassemi and N. Chen},
  pages     = {105523--105553},
  publisher = {Curran Associates, Inc.},
  title     = {ToolRL: Reward is All Tool Learning Needs},
  url       = {https://proceedings.neurips.cc/paper_files/paper/2025/file/97c5b2707228e7e3fb67e4ecc2e0e607-Paper-Conference.pdf},
  volume    = {38, Main Conference},
  year      = {2025}
}

@article{zhang2025landscape,
  title   = {The Landscape of Agentic Reinforcement Learning for {LLM}s: A Survey},
  author  = {Guibin Zhang and Hejia Geng and Xiaohang Yu and Zhenfei Yin and Zaibin Zhang and Zelin Tan and Heng Zhou and Zhong-Zhi Li and Xiangyuan Xue and Yijiang Li and Yifan Zhou and Yang Chen and Chen Zhang and Yutao Fan and Zihu Wang and Songtao Huang and Francisco Piedrahita Velez and Yue Liao and Hongru WANG and Mengyue Yang and Heng Ji and Jun Wang and Shuicheng YAN and Philip Torr and LEI BAI},
  journal = {Transactions on Machine Learning Research},
  issn    = {2835-8856},
  year    = {2026},
  url     = {https://openreview.net/forum?id=RY19y2RI1O},
  note    = {Survey Certification}
}

@article{huang2025deep,
  title   = {Deep research agents: A systematic examination and roadmap},
  author  = {Huang, Yuxuan and Chen, Yihang and Zhang, Haozheng and Li, Kang and Zhou, Huichi and Fang, Meng and Yang, Linyi and Li, Xiaoguang and Shang, Lifeng and Xu, Songcen and others},
  journal = {arXiv preprint arXiv:2506.18096},
  year    = {2025}
}

@inproceedings{li2025search,
  title     = {Search-o1: Agentic search-enhanced large reasoning models},
  author    = {Li, Xiaoxi and Dong, Guanting and Jin, Jiajie and Zhang, Yuyao and Zhou, Yujia and Zhu, Yutao and Zhang, Peitian and Dou, Zhicheng},
  booktitle = {Proceedings of the 2025 Conference on Empirical Methods in Natural Language Processing},
  pages     = {5420--5438},
  year      = {2025}
}

@inproceedings{wang2024searching,
  title     = {Searching for best practices in retrieval-augmented generation},
  author    = {Wang, Xiaohua and Wang, Zhenghua and Gao, Xuan and Zhang, Feiran and Wu, Yixin and Xu, Zhibo and Shi, Tianyuan and Wang, Zhengyuan and Li, Shizheng and Qian, Qi and others},
  booktitle = {Proceedings of the 2024 Conference on Empirical Methods in Natural Language Processing},
  pages     = {17716--17736},
  year      = {2024}
}

@inproceedings{he2026hierarchyofgroups,
  title     = {Hierarchy-of-Groups Policy Optimization for Long-Horizon Agentic Tasks},
  author    = {Shuo He and Lang Feng and Qi Wei and Xin Cheng and Lei Feng and Bo An},
  booktitle = {The Fourteenth International Conference on Learning Representations},
  year      = {2026},
  url       = {https://openreview.net/forum?id=T8Dev99qnz}
}


\appendix
\section{Derivations and Proofs}
\label{sec:proof}
\subsection{Proof of Theorem~\ref{theorem:1}}
\label{proof:1}

\paragraph{Setup.}
Recall that a trajectory is denoted by
\begin{equation}
    o = (h_0 \xrightarrow{\tau_1} h_1 \xrightarrow{\tau_2} \cdots \xrightarrow{\tau_T} h_T ),
\end{equation}
where all trajectories share a common root node $h_0=q$. Its probability under the original policy is
\begin{equation}
    \pi_\theta(o)=\prod_{t=1}^T \pi_\theta(\tau_t\mid h_{t-1}).
\end{equation}
At each expansion stage, among all currently active nodes, node $h$ is selected with probability proportional to
$\exp(\gamma \mathrm{val}(h))$, as specified in \eqref{eq:expand_prob_theorem}, and then expanded according to $\pi_\theta(\cdot\mid h)$. We further define the partial cumulative value as
\begin{equation}
    \label{eq:partial_G}
    V(o_t) := \sum_{k=1}^{t-1} \mathrm{val}(h_k),
\end{equation}
where $o_t$ denotes the partial trajectory up to turn $t$.
\paragraph{Proof strategy.}
We first derive the induced distribution over partial trajectories at each depth. We then show by induction that the depth-$t$ distribution is proportional to the original policy probability multiplied by the exponential information-gain weight. Finally, specializing to terminal prefixes yields the desired complete-trajectory distribution.

\paragraph{Step 1: One-step expansion from the root.}
At depth $0$, the root is deterministic:
\begin{equation}
    \mathbb{P}(h_0=q)=1.
\end{equation}
Since the root is always expanded, the probability of generating a depth-$1$ node $h_1=(\tau_1)$ is
\begin{equation}
    \mu_\theta^{(1)}(h_1)=\pi_\theta(\tau_1\mid h_0).
\end{equation}
We have
\begin{equation}
    \begin{aligned}
        \mu_\theta^{(1)}(o_1) & =\mu_\theta^{(1)}(h_1)                                            \\
                              & \propto \pi_\theta(\tau_1\mid h_0)\exp\bigl(\gamma \times 0\bigr) \\
                              & =\pi_\theta(\tau_1\mid h_0)\exp\bigl(\gamma V(o_1)\bigr).
    \end{aligned}
\end{equation}

\paragraph{Step 2: Induced distribution at an arbitrary layer $t\geq 2$.}
Let $\mathcal{A}_t$ denote the set of depth-$t$ partial trajectories. We now characterize the induced distribution over $\mathcal{A}_t$.
Suppose that at depth $t-1$, the empirical distribution over $\mathcal{A}_{t-1}$ converges to some limit $\mu_\theta^{(t-1)}$. Fix the stage budget at depth $t-2$ to be $B_{t-2}$. For any $o_{t-1}$, let $N(o_{t-1})$
denote the number of times that $o_{t-1}$ appears in $\mathcal{A}_{t-1}$. Then, by definition we have
\begin{equation}
    \sum_{o_{t-1}} N(o_{t-1}) = B_{t-2}.
\end{equation}
According to the rule in \eqref{eq:expand_prob_theorem}, for each independent sampling trial at depth $t-1$, the probability that prefix $o_{t-1}$ is selected for expansion is
\begin{equation}
    \mathbb{P}\bigl(\text{select } o_{t-1}\bigr)
    =
    \frac{N(o_{t-1})\exp\bigl(\gamma \,\mathrm{val}(h_{t-1})\bigr)}
    {\sum_{\tilde o_{t-1}} N(\tilde o_{t-1})\exp\bigl(\gamma \,\mathrm{val}(\tilde h)\bigr)}.
    \label{eq:select-prefix-prob}
\end{equation}
Conditioned on selecting $o_{t-1}$, the next action $\tau_t$ is sampled according to $\pi_\theta(\tau_t\mid h_{t-1})$. Therefore, for the depth-$t$ prefix $o_t=(o_{t-1} \xrightarrow{\tau_t} h_t)$,
its generation probability satisfies
\begin{equation}
    \mathbb{P}(\mathrm{generate}\, o_t)
    =
    \mathbb{P}(\mathrm{select}\, o_{t-1})\,
    \pi_\theta(\tau_t\mid h_{t-1}).
    \label{eq:generate_ot}
\end{equation}
Substituting \eqref{eq:select-prefix-prob} into \eqref{eq:generate_ot}, we obtain
\begin{equation}
    \begin{aligned}
         & \mathbb{P}(\mathrm{generate}\ o_t)                                                        \\
         & =
        \frac{N(o_{t-1})\exp\bigl(\gamma \mathrm{val}(h_{t-1})\bigr)}
        {\sum_{\tilde o_{t-1}} N(\tilde o_{t-1})\exp\bigl(\gamma \mathrm{val}(\tilde h_{t-1})\bigr)} \\
         & \quad\times\pi_\theta(\tau_t\mid h_{t-1}).
    \end{aligned}
    \label{eq:generate_ot_explicit}
\end{equation}
When $B_{t-2}\to\infty$, by the law of large numbers, for every prefix $o_{t-1}$,
\begin{equation}
    \frac{N(o_{t-1})}{B_{t-2}}
    \overset{a.s.}{\longrightarrow}
    \mu_\theta^{(t-1)}(o_{t-1}).
\end{equation}
Therefore, dividing both the numerator and denominator of \eqref{eq:generate_ot_explicit} by $B_{t-2}$ yields
\begin{equation}
    \begin{aligned}
         & \mathbb P(\mathrm{generate}\ o_t)                                       \\
         & \overset{a.s.}{\longrightarrow}
        \frac{
            \mu_\theta^{(t-1)}(o_{t-1})\exp\bigl(\gamma \mathrm{val}(h_{t-1})\bigr)
        }{
            \sum_{\tilde o_{t-1}}
            \mu_\theta^{(t-1)}(\tilde o_{t-1})
            \exp\bigl(\gamma \mathrm{val}(\tilde h_{t-1})\bigr)
        } \\
         & \quad\times\pi_\theta(\tau_t\mid h_{t-1}).
    \end{aligned}
    \label{eq:limit_generate}
\end{equation}
Hence, noting that the denominator in \eqref{eq:limit_generate} is a finite normalizing constant independent of $o_t$, the limiting depth-$t$ distribution satisfies
\begin{equation}
    \begin{aligned}
         & \mu_\theta^{(t)}(o_t) \propto  \mu_\theta^{(t-1)}(o_{t-1}) \times \\
         & \quad\exp\bigl(\gamma \mathrm{val}(h_{t-1})\bigr)
        \pi_\theta(\tau_t \mid h_{t-1}).
    \end{aligned}
    \label{eq:prefix-recursion}
\end{equation}

\paragraph{Step 3: Induction formula for prefix distributions.}
We now prove by induction that, for every depth $t\ge 1$,
\begin{equation}
    \mu_\theta^{(t)}(o_t)\propto \pi_\theta(o_t)\exp\bigl(\gamma V(o_t)\bigr),
    \label{eq:prefix-dist}
\end{equation}
where $\pi_\theta(o_t):=\prod_{k=1}^t \pi_\theta(\tau_k\mid h_{k-1})$ is the original policy probability of the prefix $o_t$.

\emph{Base case.}
The claim has already been shown.

\emph{Inductive step.}
Assume that \eqref{eq:prefix-dist} holds for depth $t-1$. Then, by substituting the inductive hypothesis into \eqref{eq:prefix-recursion} and using the transitivity of proportionality, we obtain
\begin{equation}
    \begin{aligned}
         & \mu_\theta^{(t)}(o_t)\propto\pi_\theta(o_{t-1})\exp\bigl(\gamma V(o_{t-1})\bigr) \times \\
         & \quad\exp\bigl(\gamma \mathrm{val}(h_{t-1})\bigr)\pi_\theta(\tau_t\mid h_{t-1}).
    \end{aligned}
\end{equation}
By using $\pi_\theta(o_t)=\pi_\theta(o_{t-1})\pi_\theta(\tau_t\mid h_{t-1})$ and $V(o_t)=V(o_{t-1})+\mathrm{val}(h_{t-1})$,
we obtain
\begin{equation}
    \mu_\theta^{(t)}(o_t)\propto \pi_\theta(o_t)\exp\bigl(\gamma V(o_t)\bigr),
\end{equation}
which proves the inductive step.
Therefore, \eqref{eq:prefix-dist} holds for all $t=1,\dots,T$.
\qed

\subsection{Proof of Theorem~\ref{theorem:2}}
\label{proof:2}

Following the Donsker-Varadhan variational formula~\cite{donsker1975asymptotic}, we show that the optimizer of \eqref{eq:u_variational_objective} is the induced trajectory distribution $\mu_\theta(o)\propto\pi_\theta(o)\exp\bigl(\gamma V(o)\bigr)$. We optimize over all trajectory distributions \(\mu\). Let
\begin{equation}
    J(\mu)
    :=
    \mathbb{E}_{o\sim\mu}[V(o)]
    -\frac{1}{\gamma}\mathbb{D}_{\mathrm{KL}}(\mu\|\pi_\theta).
\end{equation}
Using the definition of KL divergence, we have
\begin{equation}
    \mathbb{D}_{\mathrm{KL}}(\mu\|\pi_\theta)
    =
    \mathbb{E}_{o\sim\mu}
    \left[
        \log \frac{\mu(o)}{\pi_\theta(o)}
        \right].
\end{equation}
Hence,
\begin{equation}
    J(\mu)
    =
    \sum_o \mu(o) V(o)
    -\frac{1}{\gamma}
    \sum_o \mu(o)\log\frac{\mu(o)}{\pi_\theta(o)}.
\end{equation}
Rearranging terms gives
\begin{equation}
    \label{eq:Ju}
    J(\mu)
    =
    \frac{1}{\gamma}
    \sum_o \mu(o)
    \log\frac{\exp(\gamma V(o))\pi_\theta(o)}{\mu(o)}.
\end{equation}
Now we define the normalized distribution
\begin{equation}
    \mu^\star(o)
    :=
    \frac{\pi_\theta(o)\exp(\gamma V(o))}
    {\sum_{\tilde o} \pi_\theta(\tilde o)\exp(\gamma V(\tilde o))}.
\end{equation}
Let
\begin{equation}
    Z := \sum_{\tilde o} \pi_\theta(\tilde o)\exp(\gamma V(\tilde o))
\end{equation}
be the normalization constant. Then
\begin{equation}
    \label{eq:1}
    \pi_\theta(o)\exp(\gamma V(o))=Z\,\mu^\star(o).
\end{equation}
Substituting \eqref{eq:1} into \eqref{eq:Ju}, we obtain
\begin{equation}
    \begin{aligned}
        J(\mu) & =\frac{1}{\gamma}\sum_o \mu(o)\log\frac{Z\,\mu^\star(o)}{\mu(o)}                      \\
               & =\frac{1}{\gamma}\log Z-\frac{1}{\gamma}\sum_o \mu(o)\log\frac{\mu(o)}{\mu^\star(o)}.
    \end{aligned}
\end{equation}
Therefore,
\begin{equation}
    J(\mu)
    =
    \frac{1}{\gamma}\log Z
    -\frac{1}{\gamma}\mathbb{D}_{\mathrm{KL}}(\mu\|\mu^\star).
\end{equation}
Since KL divergence is always nonnegative, with equality if and only if \(\mu=\mu^\star\), it follows that
\begin{equation}
    J(\mu)\le \frac{1}{\gamma}\log Z,
\end{equation}
and the equality holds if and only if \(\mu=\mu^\star\). \qed

\section{Experiment Details}
\label{sec:experiment_details}

\subsection{Training Details}
\label{sec:training_details}

\begin{figure*}[t]
    \centering
    \includegraphics[width=\linewidth]{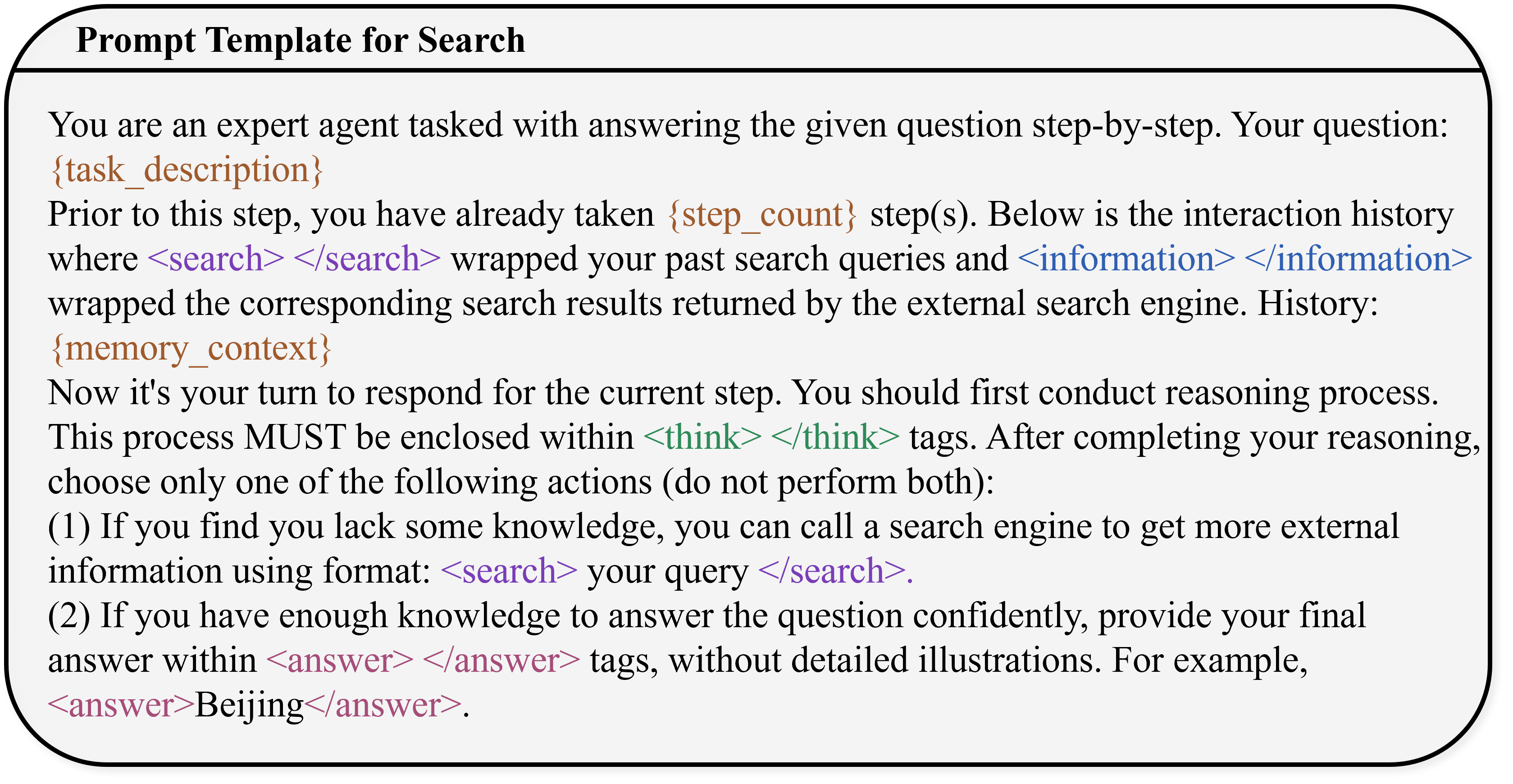}
    \caption{Prompt template for the search agent.}
    \label{fig:prompt_template}
\end{figure*}

\paragraph{Hyperparameters for Search-Augmented QA.} Our hyperparameter configuration directly follows GiGPO~\cite{feng2025group}. We set the maximum prompt length to 4096 tokens and the maximum response length to 512 tokens. The actor learning rate is set to $1\times 10^{-6}$. We use a rule-based reward function, assigning a reward of 1 to successful trajectories and 0 otherwise, while imposing a penalty of $-0.01$ for invalid actions. The training batch size is 256, with training data constructed by mixing NQ~\cite{kwiatkowski2019natural} and HotpotQA~\cite{yang2018hotpotqa} datasets. We use E5~\cite{wang2022text} as the retriever, returning the top-3 most relevant results for each query, and expose the full retrieval history to the agent. The rollout temperature and validation temperature are set to 1.0 and 0.0, respectively. The KL regularization coefficient is set to 0.001. Our IGRPO expanding temperature $\gamma$ is set to 1.0 without further tuning. Qwen2.5-3B-Instruct~\cite{yang2024qwen2} is trained on 4$\times$H20 GPUs, while Qwen2.5-7B-Instruct is trained on 8$\times$H20 GPUs.

\paragraph{Implementation Details.} We build our codebase on top of the \texttt{verl-agent} framework~\cite{feng2025group}. In particular, on top of this framework, we reproduce IGPO~\cite{wang2025information} and adapt it to use the E5 retriever in our search-augmented QA setting. Our preliminary experiments as well as the proposed IGRPO are then both developed based on this reproduced IGPO implementation. For our IGRPO, we adopt a practical strategy to promote trajectory diversity under limited computational resources: each node is expanded at most twice per stage except for the init stage. Note that under this constraint, our main experiments and ablations demonstrate IGRPO's effectiveness and support our central premise that informativeness-guided trajectory generation benefits policy learning. We also emphasize that the information gain signal is only used during training for rollout allocation; at inference time, the agent follows the learned policy without access to the ground-truth answer. For Tree-GRPO~\cite{ji2025tree}, we implement it based on its official codebase, with $m=2$, $n=3$, $k=4$, and $l=1$. The remaining baseline results reported in Table \ref{tab:main_results} are taken from GiGPO, since we use an identical training configuration. We directly use the Search-R1 datasets~\cite{jin2025search}.

\paragraph{Search Prompt.} The prompt template for the search agent is shown in Figure \ref{fig:prompt_template}. We construct this template using Python-style string formatting, where placeholders enclosed in curly braces, including {\color[HTML]{A05A2C}\texttt{\{step\_count\}}}, {\color[HTML]{A05A2C}\texttt{\{memory\_context\}}} and {\color[HTML]{A05A2C}\texttt{\{task\_description\}}}, are dynamically populated at runtime. The previous interaction history is explicitly incorporated into the prompt, where all prior search queries are sequentially inclosed by {\color[HTML]{7A3DB8}\texttt{<search></search>}} and retrieved evidence is marked by {\color[HTML]{2F5FB3}\texttt{<information></information>}}.

The agent is first required to perform explicit reasoning within {\color[HTML]{2E8B57}\texttt{<think></think>}} tags. It must then take exactly one action: either calling the search engine using {\color[HTML]{7A3DB8}\texttt{<search></search>}} when additional knowledge is required, or producing the final answer using {\color[HTML]{A64D79}\texttt{<answer></answer>}} when it is confident enough to respond. Such a formulation makes retrieval-augmented generation more interpretable and easier to standardize across search trajectories.

\subsection{Additional Results on Entropy Dynamics}
\label{sec:entropy_analysis}

\begin{figure}[t]
    \centering
    \includegraphics[width=\linewidth]{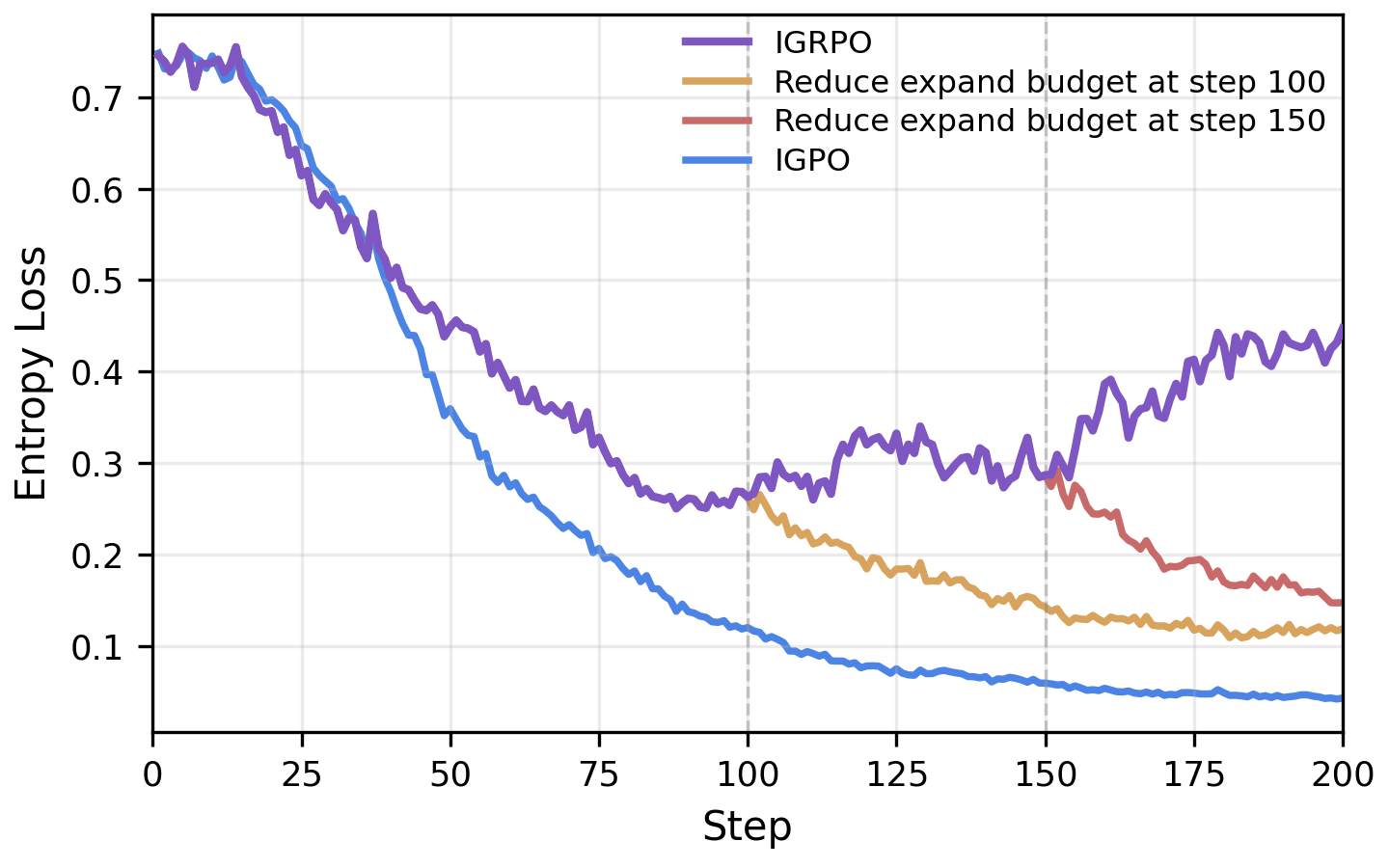}
    \caption{Entropy dynamics on the 3B backbone. Dashed lines mark reductions in the per-node expansion limit at step 100 (yellow) or 150 (red). Default IGRPO and IGPO are included for comparison.}
    \label{fig:additional_training_dynamics}
\end{figure}

We further examine how rollout allocation affects exploration by reducing the maximum number of expansions per node at particular training step.
As shown in Figure~\ref{fig:additional_training_dynamics}, both reductions lead to a decline in entropy. We also observe lower final performance relative to standard IGRPO. Together, these findings suggest that sustaining exploration during training benefits task performance.

\subsection{Additional Ablation Results.}
\label{sec:additional_ablation_study}

\begin{table}[!t]
    \centering
    \fontsize{10pt}{12pt}\selectfont
    \setlength{\tabcolsep}{1pt}
    \begin{tabular*}{\columnwidth}{@{\extracolsep{\fill}}l*{8}{c}@{}}
        \toprule
        Criterion & NQ & Tri. & Pop. & Hot. & 2Wi. & Mu. & Bam. & Avg. \\
        \midrule
        $s(\cdot)$ & \textbf{45.7} & 59.8 & 43.6 & \textbf{40.0} & \textbf{40.4} & \underline{13.7} & \underline{31.2} & \underline{39.2} \\
        $\mathrm{IG}(\cdot)$ & 45.5 & \textbf{61.6} & \underline{46.7} & 36.8 & 38.0 & 13.3 & 29.6 & 38.8 \\
        default & \underline{45.6} & \underline{60.9} & \textbf{47.5} & \underline{38.9} & \underline{40.2} & \textbf{14.1} & \textbf{32.0} & \textbf{39.9} \\
        \bottomrule
    \end{tabular*}
    \caption{Selection criterion $\mathrm{val}(\cdot)$ ablation on the 3B backbone (EM, \%). Avg. is the unweighted mean across seven datasets. $s(\cdot)$ and $\mathrm{IG}(\cdot)$ denote using only the answer score and information gain, respectively; default uses their arithmetic mean.}
    \label{tab:criterion_compare}
\end{table}

\begin{table}[!t]
    \centering
    \fontsize{10pt}{12pt}\selectfont
    \setlength{\tabcolsep}{2pt}
    \begin{tabular*}{\columnwidth}{@{\extracolsep{\fill}}lccc@{}}
        \toprule
        \makecell[l]{Reward Design} & Stage1 & Stage2 & Avg. \\
        \midrule
        All zeros & 0 & 0 & 36.9 \\
        Full score & $s_\theta(h;a)$ & $s_\theta(h;a)$ & 37.9 \\
        Half score & $0.5s_\theta(h;a)$ & $0.5s_\theta(h;a)$ & 38.4 \\
        Rescaled & $0.5s_\theta(h;a)$ & $s_\theta(h;a)$ & \textbf{39.9} \\
        Thresholded & $0.5s_\theta(h;a)$ & $\mathbb{I}[s_\theta(h;a)\geq0.5]$ & \underline{39.2} \\
        \bottomrule
    \end{tabular*}
    \caption{Reward designs for the truncated trajectories. Stage1 and Stage2 cover the first 150 and last 50 training iterations, respectively. $s_\theta(h;a)$ is the answer score at the last observed node $h$ of a truncated trajectory, and $\mathbb{I}[\cdot]$ is the indicator function. Avg. is the unweighted mean EM (\%) across seven datasets.}
    \label{tab:reward_compare}
\end{table}

\paragraph{The selection criterion design.} The selection criterion $\mathrm{val}(\cdot)$ determines how each node is valued for rollout allocation. Table~\ref{tab:criterion_compare} compares the answer score, information gain, and their arithmetic mean. The mean combines current answerability with the latest turn's incremental progress. Each criterion has advantages on different datasets, while their average EM scores remain close and all exceed the 3B baseline averages reported in Table~\ref{tab:main_results}. Motivated by the preliminary observation in Figure~\ref{fig:pre}, a plausible explanation is that the cumulative trajectory value $V(o)$ under each criterion is positively associated with the final reward $R(o)$. These results support flexibility in the choice of $\mathrm{val}(\cdot)$ and motivate future exploration of smaller scoring models tailored to a wider range of tasks.

\paragraph{The reward design.} Although our work primarily focuses on informativeness-guided node selection and the induced trajectory distribution, practical implementations must also assign appropriate rewards $R(o)$ to trajectories $o$ truncated when computational constraints prevent further expansion. We therefore design $R(o)$ specifically for these truncated trajectories, while those that terminate naturally or reach the interaction limit receive the reward returned by the environment. Notably, credit assignment remains at the trajectory level; we do not explore finer-grained schemes, as our focus is on how training trajectories are generated.

To better align learning under finite computational resources with the induced limiting teacher distribution, we estimate the potential value of each truncated trajectory $o$, rather than assigning zero reward. Our preliminary experiment in Figure~\ref{fig:pre} shows that successful trajectories exhibit higher average cumulative IG at intermediate turns, suggesting its relevance to eventual reward. We therefore base $R(o)$ on the cumulative IG at the last observed node $h$, which equals its answer score $s_\theta(h;a)$.

As shown in Table~\ref{tab:reward_compare}, we compare several reward designs for truncated trajectories. \texttt{All zeros} yields the lowest average EM, suggesting the importance of estimating the potential value. \texttt{Rescaled} performs best; one possible explanation is that the $0.5$ coefficient reduces the risk of over-rewarding incomplete trajectories early in training (acting as a penalty for trajectories not expanded further), while the full score later provides stronger feedback for informative partial trajectories (as the model has by then acquired much search capability).

\section{Workflow}
\label{sec:algorithm}

\begin{algorithm}[!t]
    \caption{IGRPO}
    \label{alg:igrpo}
    \begin{algorithmic}[1]
        \Statex \textbf{Require:} Policies $\pi_{\theta}$, $\pi_{\mathrm{ref}}$; data $\mathcal{D}$.
        \Statex Parameters $\gamma,\beta,\epsilon$; budgets $\{B_i\}$; turn limit $T$.
        \For{each training iteration}
        \State $\theta_{\mathrm{old}} \gets \theta$.
        \State Sample a batch of $(q,a)\sim\mathcal{D}$.
        \State Initialize root $h_0$ for each $q$.
        \State $\mathcal{A}_0 \gets \{h_0\}$; $\mathcal{O}\gets\emptyset$.

        \Statex \Comment{IG-guided rollout}
        \For{$i = 0$ to $T-1$}
        \If{$\mathcal{A}_i=\emptyset$}
        \State \textbf{break}
        \EndIf
        \Statex \Comment{Batched answer score computation}
        \For{$h\in\mathcal{A}_i$}
        \State Compute $s_\theta(h;a)$.
        \State Compute $\mathrm{IG}(h)$ and $\mathrm{val}(h)$.
        \EndFor
        \State Compute $p_i(\cdot)$ within each group.
        \State $\mathcal{A}_{i+1}\gets\emptyset$.
        \For{$b = 1$ to $B_i$}
        \State Sample $h\sim p_i(\cdot)$ per group.
        \State Sample $\tau\sim \pi_{\theta_{\mathrm{old}}}(\cdot\mid h)$.
        \State Execute $\tau$ to obtain child $h'$.
        \If{$h'$ is terminal}
        \State Add the path to $h'$ to $\mathcal{O}$.
        \Else
        \State Add $h'$ to $\mathcal{A}_{i+1}$.
        \EndIf
        \EndFor
        \For{unexpanded $h\in\mathcal{A}_i$}
        \State Form path $o$ ending at $h$.
        \State Compute $R(o)$; add $o$ to $\mathcal{O}$.
        \EndFor
        \EndFor
        \For{$h\in\mathcal{A}_T$}
        \State Add the path to $h$ to $\mathcal{O}$.
        \EndFor

        \Statex \Comment{Policy update}
        \State Collect $\{o_i\}$ from $\mathcal{O}$.
        \State Update $\theta$ by maximizing $\mathcal{J}_{\mathrm{IGRPO}}(\theta)$.
        \EndFor
    \end{algorithmic}
\end{algorithm}

Algorithm~\ref{alg:igrpo} shows the overall workflow of IGRPO. Instead of directly sampling complete trajectories from the policy, IGRPO constructs paths through a tree-structured rollout process, where active nodes are expanded layer by layer under information gain-based allocation.
This procedure approximates the teacher distribution $\mu_{\theta_{\mathrm{old}}}$ over trajectories, based on which policy optimization is performed.

\section{Case Study}
\label{sec:case_study}

\begin{figure}[t]
    \centering
    \includegraphics[width=\linewidth]{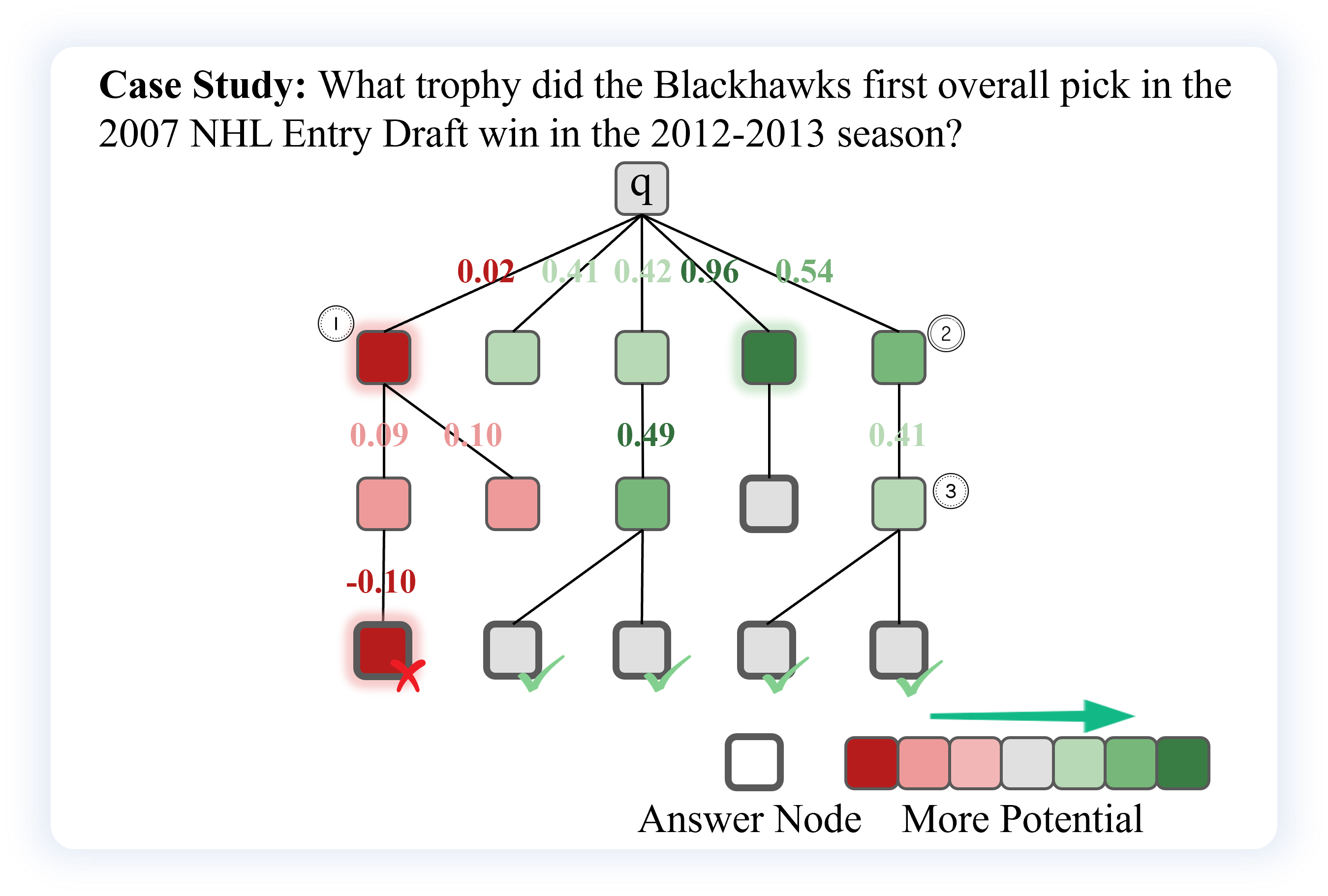}
    \caption{Case study of the tree-structured rollout behaviors of IGRPO. The number on each edge indicates the information gain obtained by the corresponding step. The three numbered nodes are further analyzed in detail. }
    \label{fig:case_study_overall}
\end{figure}

\begin{figure}[!t]
    \centering
    \includegraphics[width=\linewidth]{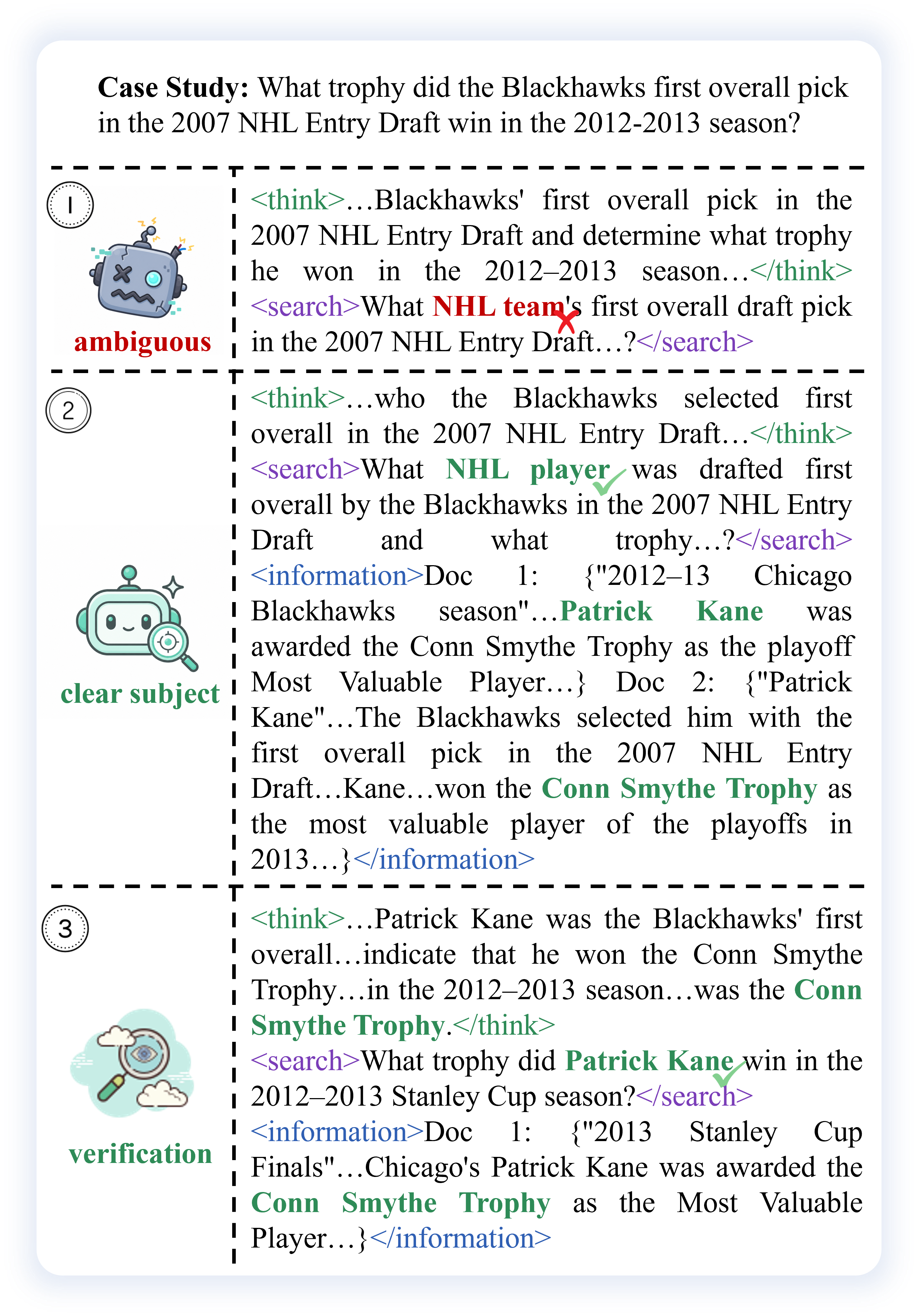}
    \caption{Detailed analysis of the three highlighted nodes in the case study. Node 1 issues an ambiguous query and fails to target the key entity. Node 2 uses the correct subject and retrieves useful evidence, while its child, Node 3, further verifies the answer and confirms the Conn Smythe Trophy.}
    \label{fig:case_study_detail}
\end{figure}

Figure~\ref{fig:case_study_overall} presents the overall tree structure of a representative example from the multi-hop QA benchmark HotpotQA~\cite{yang2018hotpotqa}. The rollout tree shows that IGRPO induces a clear bias toward branches with larger informativeness value, so that the final sampled trajectories are concentrated on more informative search paths. At the same time, the allocation remains probabilistic rather than deterministic: nodes with lower informativeness value can still be expanded at each stage, which preserves exploration and avoids overly greedy rollout behavior. Moreover, the transition from $\mathcal{A}_2$ to $\mathcal{A}_3$ clearly illustrates how low informative nodes are gradually filtered out during rollout. These nodes continue to receive less expansion budget over time and ultimately fail to produce correct answers, whereas more informative branches are retained and expanded into successful trajectories.

As shown in Figure~\ref{fig:case_study_detail}, we further detail the three highlighted nodes in Figure~\ref{fig:case_study_overall}. Node 1 issues an ambiguous query and obtains misleading search evidence, causing its subsequent continuations to fail to reach the correct answer. In contrast, Node 2 and its child Node 3 progressively move toward the correct prediction. Notably, the search results of Node 2 already contain the correct answer, while Node 3 further verifies it before producing the final decision. Such verification behavior frequently emerges during training across different models, further highlighting the importance of allocating rollout budget to informative nodes.


\end{document}